\documentclass[manuscript,authorversion,nonacm]{acmart}
\usepackage{subcaption}
\usepackage{hyperref}
\usepackage{algorithm}
\usepackage{algpseudocode}
\usepackage{tikz,pgf}
\usepackage{epsfig,amsmath,ifthen,comment}
\usepackage{color}
\usepackage{bm}
\usepackage{geometry}
\usepackage[capitalize]{cleveref}
\usepackage{soul}

\usepackage[normalem]{ulem}
\newcommand{\stkout}[1]{\ifmmode\text{\sout{\ensuremath{#1}}}\else\sout{#1}\fi}

\newtheorem{dfn}{Definition}
\def\ci{\perp\!\!\!\perp}
\DeclareMathOperator{\E}{\mathbb{E}}

\DeclareMathOperator{\doo}{do}

\AtBeginDocument{%
  \providecommand\BibTeX{{%
    \normalfont B\kern-0.5em{\scshape i\kern-0.25em b}\kern-0.8em\TeX}}}

\setcopyright{rightsretained}
\acmYear{2023}

\acmJournal{CSUR}

\def\eg{\emph{e.g., }} 
\def\ie{\emph{i.e., }} 

\begin{document}


\title{An Introduction to Causal Inference Methods \\for Observational Human-Robot Interaction Research}
\author{Jaron J.R. Lee}
\email{jaron.lee@jhu.edu}
\orcid{}
\author{Gopika Ajaykumar}
\email{gopika@cs.jhu.edu}
\author{Ilya Shpitser}
\email{ilyas@cs.jhu.edu}
\author{Chien-Ming Huang}
\email{chienming.huang@jhu.edu}
\affiliation{%
  \institution{Johns Hopkins University}
  \streetaddress{3400 North Charles Street}
  \city{Baltimore}
  \state{Maryland}
  \postcode{21218}
  \country{USA}
}

%
%
%
%
%
%

\renewcommand{\shortauthors}{Lee et al.}

\begin{abstract}

    Quantitative methods in Human-Robot Interaction (HRI) research have primarily relied upon randomized, controlled experiments in laboratory settings. However, such experiments are not always feasible when external validity, ethical constraints, and ease of data collection are of concern. Furthermore, as consumer robots become increasingly available, increasing amounts of real-world data will be available to HRI researchers, which prompts the need for quantative approaches tailored to the analysis of observational data.  In this article, we present an alternate approach towards quantitative research for HRI researchers using methods from causal inference that can enable researchers to identify causal relationships in observational settings where randomized, controlled experiments cannot be run. We highlight different scenarios that HRI research with consumer household robots may involve to contextualize how methods from causal inference can be applied to observational HRI research. 
    We then provide a tutorial summarizing key concepts from causal inference using a graphical model perspective and link to code examples throughout the article, which are available at \url{https://gitlab.com/causal/causal_hri}. Our work paves the way for further discussion on new approaches towards observational HRI research while providing a starting point for HRI researchers to add causal inference techniques to their analytical toolbox.
\end{abstract}

\ccsdesc[500]{Human-centered computing~Field studies}
\ccsdesc[300]{Human-centered computing~Laboratory experiments}

\keywords{Causal Inference, Observational Research, Longitudinal Studies, Quantitative Methods, Human-Robot Interaction}

\maketitle
\section{Introduction}\label{sec:intro}
\begin{figure}[ht]
\includegraphics[width=\textwidth]{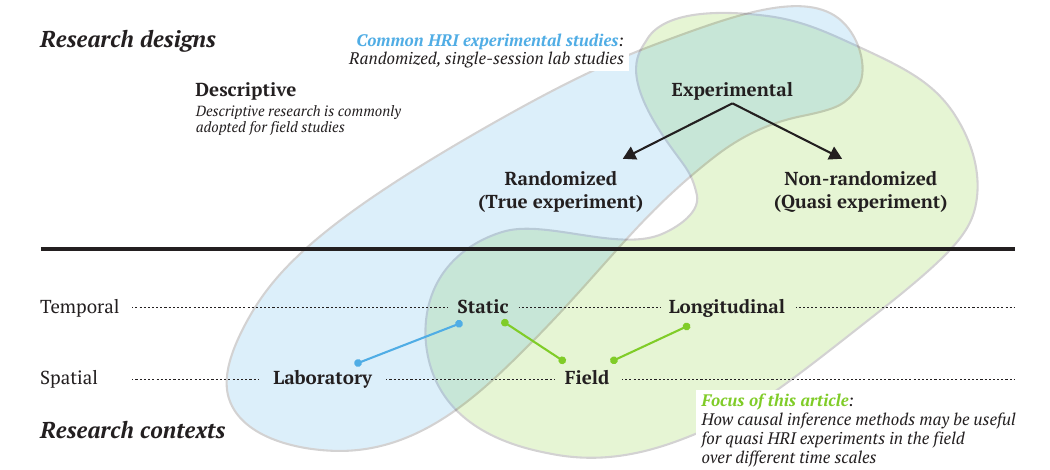}
\caption{An overview of commong HRI research methods. Each research method is related to a particular research design (top panel) and research context (bottom panel). HRI studies commonly involve randomized, laboratory-based methods (\emph{blue}). Our paper focuses on field-based methods that can be applied over different time scales, which may help HRI researchers conduct experiments with greater external validity while determining causal relationships without the use of randomization (\emph{green}).} 
\label{fig:taxonomy}
\end{figure}

Human-Robot Interaction (HRI) research has involved a variety of methodological approaches toward evaluating and understanding the impact of interactive robotic systems and interaction techniques (Figure \ref{fig:taxonomy}). Within the diversity of research methodologies, randomized, laboratory-based methods have emerged as one of the most common approaches for studying human-robot interaction \cite{hoffmanPrimerConductingExperiments2020,bethelQualitativeInterviewTechniques2020}. Laboratory experiments allow experimenters to manipulate one or more independent variables (\eg robot height) to observe their effect on one or more dependent variables (\eg robot persuasiveness). Thanks to the controlled nature of laboratory environments, experimenters are better able to control variables of interest, perform random assignment of participants to control and experimental groups, and infer causal relations \cite{hoffmanPrimerConductingExperiments2020}.

However, laboratory-based experiments can provide a limited view of how a robotic system or human-robot interaction will operate in the real-world.  For research involving robots intended for use in an application domain, such as a factory (\eg \cite{guerin2015framework}), hospital (\eg \cite{hebesberger2017long}), or home (\eg \cite{scassellatiImprovingSocialSkills2018}), it can be useful to conduct a field study at the domain of interest to observe contextually relevant interactions and understand how the robotic system will fit into users' existing workflows. Studies conducted outside of laboratory settings can also enable experimenters to better observe unexpected user behaviors and interaction failures (\eg \cite{andrist2017went}) and to better understand context-dependent measures such as trust in robots (\eg \cite{hancockMetaAnalysisFactorsAffecting2011, flookImpactDifferentTypes2019}). Therefore, there has been growing interest in conducting ``in the wild'' evaluations in naturalistic settings such as schools and homes to better capture natural, emergent user behaviors (\eg \cite{jorgensen2021soft, bjorling2020can}). 

Nevertheless, real-world field studies involve several challenges, such as the difficulty of controlling the study environment (\eg \cite{van2020designing}) or special considerations involving vulnerable participant populations, that can make an experimental approach that uses randomization infeasible or unethical. Therefore, researchers may not be able to randomize participant assignment, control for all confounders, or determine the composition of study populations. As a result, most field studies have largely focused on descriptive research that uses observational and qualitative approaches and behavioral measures, often involving video analysis, interviews, questionnaires, design probes, and direct observation to understand participants’ interaction patterns and perceptions (\eg \cite{leite2013social, bjorling2020can}) (Figure \ref{fig:taxonomy}). Alternately, HRI researchers have also used quasi experiments that seek to study hypotheses without randomizing participants (\eg \cite{kidd2008robots, van2020designing}) (Figure \ref{fig:taxonomy}). Although these methods can provide valuable insight into user behaviors without requiring the use of controlled studies, they are often time-consuming due to the degree of manual annotation they require. Furthermore, they cannot be used to determine correlation or causation due to the lack of randomization. Methods from causal inference can help complement existing approaches to field studies by enabling researchers to determine causal relationships without relying on randomization.

Furthermore, we argue that new investigative tools such as causal inference will be required as robots become more commonplace in society. Indeed, such a trend can be observed in the broader field of Human-Computer Interaction (HCI). As computers became commonplace in homes and workplaces, user interaction and experience (UI/UX) design has moved from academic research, to industrial research, and finally to commercial products  \cite{myersBriefHistoryHumancomputer1998}, leading to a proliferation of designs. To evaluate and compare different designs, researchers and engineers have leveraged controlled experimental methods,  such as A/B testing, weblabs, live traffic experiments, flights, and bucket tests \cite{dengImprovingSensitivityOnline2013}, to causally establish which designs to adopt. However, conducting controlled experiments at a large scale has limited feasibility due to constraints on manpower, money, and time \cite{guptaTopChallengesFirst2019}. Furthermore, structural constraints can limit the hypotheses that can be tested (\eg interference, treatment compliance \cite{kuang2020causal}). Consequently, UI/UX researchers have begun to move toward observational causal inference methods to address these shortcomings in randomized experimentation \cite{UsingCausalInference2019,OcelotScalingObservational,dengWhenVoiceAssistant2021}. Instead of conducting a randomized experiment, they instead emulate it by applying causal inference algorithms to non-randomized data. This approach has the advantage of enabling research insights to be made on the basis of existing data---for example, the impact of changes to existing UX/UI designs or image layout algorithms can be estimated before they are deployed. 

We believe that a similar trend toward increasing amounts of observational data is happening in HRI, as it has in other fields (\eg social sciences, healthcare, economics). Household robots such as Amazon Astro are becoming commercially available and more accessible, prompting researchers to determine feasible experimental methods to optimize robot behaviors in uncontrolled settings \cite{leeDevelopingAutonomousBehaviors2023}. The growth of consumer robots represents new opportunities for HRI research, as it will become possible to collect data at a larger scale. Analytical tools incorporating causal inference can help the HRI community in leveraging this growth in field-based human-robot interaction data.

In this work, we present a set of \emph{causal inference} methods to demonstrate 
the possibilities that exist for doing hypothesis-based science in the absence of fully randomized experiments; these methods offer a new set of analytical tools that can be used to help identify and estimate causal relationships in single-session and longitudinal field studies (Figure \ref{fig:taxonomy}) and broaden the traditional perspective on experimental design in the HRI community.
These methods include graphical models \cite{pearlCausality2009}, causal identification theory \cite{tianGeneralIdentificationCondition2002,shpitserIdentificationJointInterventional2006}, adjusting for confounding, transporting inferences between domains \cite{bareinboimTransportabilityCausalEffects2012}, dealing with measurement error \cite{kurokiMeasurementBiasEffect2014}, and causal inference in a longitudinal setting \cite{hernanCausalInferenceWhat2020}. Causal inference has seen widespread application in observational social sciences \cite{fosterCausalInferenceDevelopmental2010,morganCounterfactualsCausalInference2015}, epidemiology \cite{rothmanCausationCausalInference2005}, medicine \cite{hernanUsingBigData2016,hernanEstimatingCausalEffects2006}, and other fields where controlled experiments can be technically difficult, unethical, or prohibitively expensive to conduct.  We hope that causal inference methods can similarly contribute toward answering many real-world questions in HRI and other behavioral sciences with similar aims.

We begin the paper by describing examples of scenarios involving observational HRI research with consumer robots and highlighting tools from causal inference relevant to each scenario, which we then detail further in the following sections. 
In \cref{sec:background} we provide some basic details about causal graphical models and some basic concepts from causal inference.
In Section \ref{sec:field-studies}, we study what happens when the key assumption of random assignment fails to hold in a non-longitudinal setting. In Section \ref{sec:longitudinal}, we consider simple longitudinal settings to understand what happens when we allow repeated interventions over time. In Section \ref{sec:pipeline}, we provide some patterns of analysis intended to guide would-be practitioners on how to apply the discussed methods. Finally, we conclude with a brief discussion on the benefits and limitations of causal inference and suggest some avenues for future work in Section \ref{sec:conclusion}. For Sections \ref{sec:field-studies} to \ref{sec:pipeline}, we provide Python notebooks (\url{https://gitlab.com/causal/causal_hri}) containing code examples showcasing how to implement concepts from each section.

\section{Example Use Cases of Causal Inference in Observational HRI Research}
\label{sec:hri_contexts}
In this section, we showcase different application scenarios for causal inference methods using the growing availability of household consumer robots as a motivating example.

\subsection{Handling Measured and Unmeasured Confounding}
Suppose that a company wants to introduce a ``following'' feature into their household robots where the robot follows the user into each room in their home. The company wanted to avoid forcing the feature on all users since some users may not want the robot to follow them due to privacy concerns, so they instead provides users with the option to opt in or out of adopting the feature according to their preferences. The company wants to test how the robot behavior including the feature compares to the robot behavior without the feature in eliciting user engagement.
However, since the feature was not randomly deployed to users but rather users had the choice to opt in, there may be some additional factors besides the feature itself, such as trust in robots, that may also influence user engagement. For example, users who opted to adopt the feature may have higher trust in robots, which may cause them to have higher engagement with the robot regardless of the robot's behavior. Therefore, the company would not be able to test the true effect of the robot behavior on user engagement without accounting for these additional factors. This example highlights the possibility of \emph{confounding} occurring in studies, where confounding variables (\eg trust in robots) can affect other variables (\eg robot behavior, user engagement), causing a spurious association between the other variables. Confounding is a common issue in non-randomized studies where participants can choose which study condition to participate in. 
We highlight the consequences of failing to consider confounding in non-randomized studies from a causal perspective and provide causal inference methods for addressing issues of confounding in Section \ref{sec:observed-common-causes}.

In practice, it might be difficult or impossible to collect data on all confounders -- for instance, variables such as trust cannot be measured directly, and users may be unwilling to reveal such information. In such cases, unmeasured confounding can adversely impact inference, as demonstrated in \cref{sec:unobserved_confounders}.

\subsection{Transporting Inferences Between Contexts}
Suppose that the company has decided to try rolling out the ``following'' feature among their employees as a first step before deploying the feature to all users. Compared to the general population, the company's employees are likely to be better educated, less diverse, and more comfortable with technology in society. While the new ``following'' feature may be rated well internally, company executives are concerned about how it will be received once the feature goes public, given the heterogeneity observed between the two populations.  \cref{sec:transportability} introduces methods to address transporting inferences from one domain to another.

\subsection{Managing Measurement Error}

Suppose that the company acknowledges that trust in robots is an important confounder that influences both the initial uptake and also the continued engagement, and intends to capture the said trust to inform their decisions in research and development. To measure trust in robots, the company introduces a survey aiming to probe user trust and interaction experience. However, the survey is not a perfect proxy of trust, and is at best a noisy and imperfect measurement. Although trust in robots is the true confounder, the company can only adjust for the imperfect proxy as measured by the survey. In \cref{sec:measurement_bias} we discuss methods and frameworks to think about the impact of measurement error on causal inferences.

\subsection{Seeking Causal Discovery}
Suppose that the company managed to roll out the ``following'' feature, and is seeking to discover new ways in which user engagement can be improved. The company collects a vast amount of data about how users interact with the robot, but also about all aspects about the robot behavior policy, how it was marketed, and who is purchasing the robots and where. To understand the big picture, the company might want to understand how all these factors interact with each other, and ultimately their impact on engagement, so that further improvements to company operations can be made. That is, we suspect that there are confounders, but we do not know where they lie in our problem. One way to do this is through causal discovery, which learns the causal relationships between variables from data and domain knowledge. We briefly discuss such methods in \cref{sec:causal-discovery}.

\subsection{Making Longitudinal Inference}
Suppose that the company has successfully deployed a version of the ``following'' behavior policy, but now wants to improve it. The company has found that leaving the ``following'' policy on permanently leads to decreasing engagement over time, as users start to view the robot as clingy and overly attached. The company wants to use data where users were able to freely opt in or out of the policy, to determine an improved version of the ``following'' policy that adapts over time in reaction to user engagement. This new policy is longitudinal, in the sense that there are repeated actions and reactions between the robot and user over time. We discuss methods and ideas for predicting the performance of such policies without deploying them in \cref{sec:longitudinal}.

\section{Background: Causal Graphical Models} \label{sec:background}

In this section, we introduce the core concepts of causal inference. We do this by first introducing the causal graphical model, which serves as an analytical tool. Then, we reexamine the between-subjects experimental design under the lens of causal graphical models to expose the assumptions that are made when we draw causal conclusions from randomized data. 


\subsection{Causal Inference through Graphical Models}
\label{sec:graphical-models}

For purposes of exposition, we will consider an example loosely inspired by \cite{gombolayRoboticAssistanceCoordination2016}, in which we are interested in investigating relationship between trust in the type of decision support system-- computer-issued, or robot-issued -- amongst medical practitioners.

A widely used tool in causal inference is the \emph{causal graphical model} \cite{pearlCausalDiagramsEmpirical1995, pearlCausality2009}. Causal graphical models provide a way to visualize a graph in which variables are represented as nodes, and causal relations are represented as edges between these nodes. We first consider \emph{causal directed acyclic graphs (DAGs)}, in which  all present edges are directed edges (\ie $\to$). A causal DAG represents our understanding of the causal relationships between variables in a particular problem.

For instance, if we denote the type of decision support as $A$ and subject trust as $Y$, then a lack of an edge $A \to Y$ denotes no direct causation between these variables, but this does not rule out correlation or causation through other variables. The presence of an edge $X \to Y$ denotes direct causation from $X$ to $Y$.

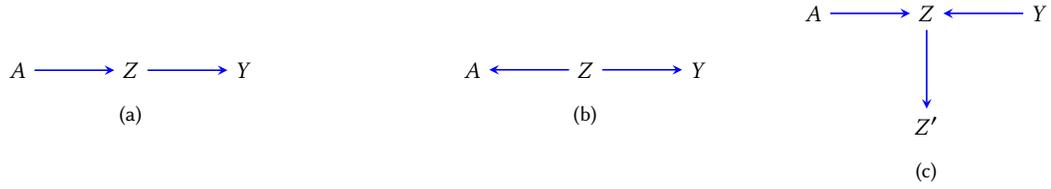
\begin{figure}[!htb]
    \begin{minipage}[l]{0.3\textwidth}
      \centering
      \begin{tikzpicture}[
          > = stealth, 
          auto,
          node distance = 1.5cm, 
          semithick 
          ]

          \tikzstyle{state}=[
          draw = none,
          fill = white,
          minimum size = 2mm
          ]
          \tikzstyle{fixed}=[
          draw=black,
          rectangle, 
          thick,
          fill = white,
          minimum size = 2mm
          ]

          \node[state] (X) {$A$};
          \node[state] (Z) [right of=X] {$Z$};
          \node[state] (Y) [right of=Z] {$Y$};

          \path[->, blue] (X) edge node {} (Z);
          \path[->, blue] (Z) edge node {} (Y);
      \end{tikzpicture}
      \subcaption{}
    \end{minipage}\hfill%
    \begin{minipage}[l]{0.3\textwidth}
      \centering
      \begin{tikzpicture}[
          > = stealth, 
          auto,
          node distance = 1.5cm, 
          semithick 
          ]

          \tikzstyle{state}=[
          draw = none,
          fill = white,
          minimum size = 2mm
          ]
          \tikzstyle{fixed}=[
          draw=black,
          rectangle, 
          thick,
          fill = white,
          minimum size = 2mm
          ]
          \node[state] (X) {$A$};
          \node[state] (Z) [right of=X] {$Z$};
          \node[state] (Y) [right of=Z] {$Y$};

          \path[->, blue] (Z) edge node {} (X);
          \path[->, blue] (Z) edge node {} (Y);

      \end{tikzpicture}
      \subcaption{} 
      \end{minipage}%
      \begin{minipage}[l]{0.3\textwidth}
          \centering
          \begin{tikzpicture}[
              > = stealth, 
              auto,
              node distance = 1.5cm, 
              semithick 
              ]

              \tikzstyle{state}=[
              draw = none,
              fill = white,
              minimum size = 2mm
              ]
              \tikzstyle{fixed}=[
              draw=black,
              rectangle, 
              thick,
              fill = white,
              minimum size = 2mm
              ]
              \node[state] (X) {$A$};
              \node[state] (Z) [right of=X] {$Z$};
              \node[state] (Zp) [below of=Z] {$Z'$};
              \node[state] (Y) [right of=Z] {$Y$};

              \path[->, blue] (X) edge node {} (Z);
              \path[->, blue] (Y) edge node {} (Z);
              \path[->, blue] (Z) edge node {} (Zp);

          \end{tikzpicture}
          \subcaption{} 
      \end{minipage}
      \caption{D-separation triplets, where $A$ denotes the type of decision support, and $Y$ denotes subject trust, as per \cite{gombolayRoboticAssistanceCoordination2016}. (a) is a chain, in which a plausible $Z$ might be a mediator such as the subject's experience working together with the robot; (b) is a fork, in which we imagine that the medical practitioners are able to choose their own decision support system, and so a plausible $Z$ could be the subject's prior experiences with robots; and (c) is a collider, in which $Z$ could be the overall efficiency of the hospital unit, in which the decision support type and the resulting subject trust are both contributing factors, and the descendant $Z'$ could be overall revenue generated by the hospital.}
      \label{fig:d-sep}
  \end{figure}

A \emph{path} is a sequence of arrows between any two variables. Every path in a DAG longer than a single edge can be thought of as a sequence of `triplets'---by inspecting every trio of variables along the path, we can deduce whether the path carries causal influence, association, or nothing at all. This logic is encoded in the \emph{d-separation} rules. D-separation works  by inspecting three types of triplets: \emph{forks}, \emph{chains}, and \emph{colliders}, as depicted in Fig. \ref{fig:d-sep}.

\begin{figure}[!htb]
          \centering
          \begin{tikzpicture}[
              > = stealth, 
              auto,
              node distance = 1.5cm, 
              semithick 
              ]

              \tikzstyle{state}=[
              draw = none,
              fill = white,
              minimum size = 2mm
              ]
              \tikzstyle{fixed}=[
              draw=black,
              rectangle, 
              thick,
              fill = white,
              minimum size = 2mm
              ]
              \node[state] (A) {$A$};
              \node[state] (B) [right of=A] {$B$};
              \node[state] (C) [above of=B] {$C$};
              \node[state] (D) [right of=B] {$D$};
              \node[state] (E) [below of=B] {$E$};
              \node[state] (F) [right of=D] {$F$};

              \path[->, blue] (A) edge node {} (B);
              \path[->, blue] (C) edge node {} (A);
              \path[->, blue] (B) edge node {} (E);
              \path[->, blue] (D) edge node {} (E);
              \path[->, blue] (E) edge node {} (F);
              \path[->, blue] (C) edge node {} (F);
              \path[->, blue] (C) edge node {} (D);

          \end{tikzpicture}
      \caption{D-separation example. The reader can check that $A \ci D \mid C$, since $A \leftarrow C \to D$ is the only path without unconditioned colliders. However, $A \not \ci D \mid C,F$ since $F$ is a descendant of $E$, and conditioning on a descendant of a collider opens up a collider path. Additionally, there is no conditional independence between $B$ and $F$ -- while we can condition on $A,C$ to block the fork, conditioning on $E$ opens the collider at $E$, while leaving it unconditioned leaves the path $B \to E \to F$ open.}
      \label{fig:d-sep-example}
  \end{figure}
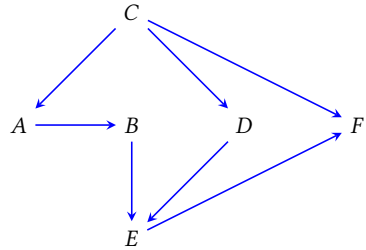

In a chain and a fork, the decision support type $A$ and trust $Y$ are marginally dependent, but conditionally independent given $Z$. In the chain, $Z$ could be a mediator such as the productivity of the work experience, while in the the fork, $Z$ could be a confounder such as prior experience working with robots. 
In a collider, the decision support type $A$ and trust $Y$ are marginally independent, but conditionally dependent given the overall efficiency of the hospital unit $Z$ or any descendant of $Z$ (such as the overall revenue generated by the hospital $Z'$). By applying these three rules for all triplets on all paths between two sets of variables given a third set, conditional independencies asserted by the model can be checked. A causal path from $A$ to $Y$ would exist if every edge along the path between $A$ and $Y$ points towards $Y$---that is, the path consists only of chain triplets. A path of association exists if there are fork and chain triplets. Finally, no association exists if there are colliders along the path, unless the collider or descendants of the collider are conditioned upon. Fig. \ref{fig:d-sep-example} provides an example illustrating the application of D-separation. 

When we reason about causal scenarios, we are often interested in \emph{counterfactuals} (also known as potential outcomes), which are variables under a hypothetical intervention that may or may not have happened in reality. For example, we can define a counterfactual variable ``subject trust where the decision support type was set to robot for all participants, regardless of their factual assignment''. This would be denoted as $Y(A=\textrm{robot})$. Counterfactuals provide a way for analysts to reason abstractly about variables under hypothetical interventions, which is useful in causal inference.

\begin{figure}[!htb]
    \begin{minipage}[l]{0.3\textwidth}
        \centering
        \begin{tikzpicture}[
            > = stealth, 
            auto,
            node distance = 1.5cm, 
            semithick 
            ]

            \tikzstyle{state}=[
            draw = none,
            fill = white,
            minimum size = 2mm
            ]
            \tikzstyle{fixed}=[
            draw=black,
            rectangle, 
            thick,
            fill = white,
            minimum size = 2mm
            ]

            \node[fixed] (X) {$a$};
            \node[state] (XX) [left of=X, xshift=1cm] {$A$};
            \node[state] (Z) [right of=X] {$Z(a)$};
            \node[state] (Y) [right of=Z] {$Y(a)$};

            \path[->, blue] (X) edge node {} (Z);
            \path[->, blue] (Z) edge node {} (Y);
        \end{tikzpicture}
        \subcaption{}
    \end{minipage}\hfill%
    \begin{minipage}[l]{0.3\textwidth}
        \centering
        \begin{tikzpicture}[
            > = stealth, 
            auto,
            node distance = 1.5cm, 
            semithick 
            ]

            \tikzstyle{state}=[
            draw = none,
            fill = white,
            minimum size = 2mm
            ]
            \tikzstyle{fixed}=[
            draw=black,
            rectangle, 
            thick,
            fill = white,
            minimum size = 2mm
            ]
            \node[fixed] (X) {$a$};
            \node[state] (XX) [left of=X, xshift=1cm] {$A$};
            \node[state] (Z) [right of=X] {$Z$};
            \node[state] (Y) [right of=Z] {$Y$};

            \path[->, blue] (Z) edge node {} (Y);

        \end{tikzpicture}
        \subcaption{} 
    \end{minipage}%
    \begin{minipage}[l]{0.3\textwidth}
        \centering
        \begin{tikzpicture}[
            > = stealth, 
            auto,
            node distance = 1.5cm, 
            semithick 
            ]

            \tikzstyle{state}=[
            draw = none,
            fill = white,
            minimum size = 2mm
            ]
            \tikzstyle{fixed}=[
            draw=black,
            rectangle, 
            thick,
            fill = white,
            minimum size = 2mm
            ]
            \node[fixed] (X) {$a$};
            \node[state] (XX) [left of=X, xshift=1cm] {$A$};
            \node[state] (Z) [right of=X] {$Z(a)$};
            \node[state] (Zp) [below of=Z] {$Z'(a)$};
            \node[state] (Y) [right of=Z] {$Y$};

            \path[->, blue] (X) edge node {} (Z);
            \path[->, blue] (Y) edge node {} (Z);
            \path[->, blue] (Z) edge node {} (Zp);

        \end{tikzpicture}
        \subcaption{} 
    \end{minipage}
    \caption{D-separation triplets from \cref{fig:d-sep} under the intervention $A=a$. where (a) is a chain, in which $Z(a)$ and $Y(a)$ represent subject experience and subject trust had the decision support system been set, possibly contrary to fact, to $a$; (b) is a fork, and so $Z$ and $Y$ retain their original meanings; and (c) is a collider, in which $Z(a)$ and $Z'(a)$ represent hospital efficiency and revenue under the hypothetical scenario decision support system assignment $a$. Notice that descendants of the variable under intervention are denoted with parentheses containing the value set under intervention.}
    \label{fig:d-sep-swig}
\end{figure}
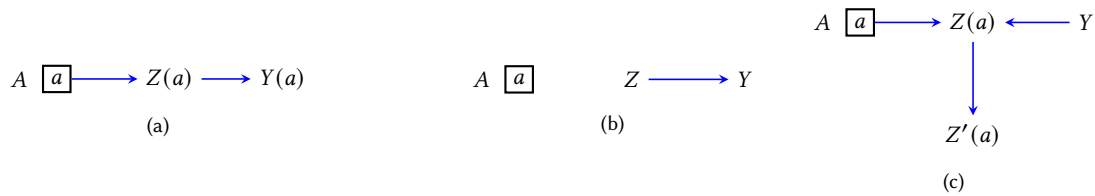

We can represent these counterfactuals using a modification of causal DAGs called \emph{Single World Interventional Graphs (SWIGs)}, which allows us to represent counterfactuals directly on the graph. Constructing a SWIG involves two steps. First, for each variable $A$ set under intervention to value $A=a$ (where $a$ here may take values ``computer'' or ``robot''), we split the original node $A$ into two parts. The first part is $A$, the original random variable, which inherits all incoming edges. The second part is the intervention value $A$, which inherits all outgoing edges. Conceptually, this represents the situation where nature generates the value of $A$, but at the last minute, the experimenter swaps it out for $A$. Second, we relabel all descendants $D$ of $A$ in the original graph to carry the intervention $D(a)$. This reminds the analyst that the variables are not the original observed variables, but rather counterfactual variables under a hypothetical scenario. To demonstrate, we consider the d-separation triplets under a hypothetical intervention of $A=a$ in Figure \ref{fig:d-sep-swig}.

SWIGs are particularly useful because potential outcomes represent what the variable would be had we performed a hypothetical intervention---exactly the question that we seek to answer with causal inference.


\subsection{Using Causal Graphical Models to Describe Randomized Experiments}\label{sec:randomized-experiments}
\begin{figure}[h]
    \begin{minipage}[l]{0.3\textwidth}
    \centering
    \begin{tikzpicture}[
        > = stealth, 
        auto,
        node distance = 1.5cm, 
        semithick 
        ]

        \tikzstyle{state}=[
        draw = none,
        fill = white,
        minimum size = 2mm
        ]
        \tikzstyle{fixed}=[
        draw=black,
        rectangle, 
        thick,
        fill = white,
        minimum size = 2mm
        ]

        \node[state] (A) {$A$};
        \node[state] (Y) [right of=A] {$Y$};

        \path[->, blue] (A) edge node {} (Y);
    \end{tikzpicture}
    \subcaption{} 
    \label{fig:rct-no-cov-1}%
    \end{minipage}%
    \begin{minipage}[l]{0.3\textwidth}
    \begin{tikzpicture}[
        > = stealth, 
        auto,
        node distance = 1.5cm, 
        semithick 
        ]

        \tikzstyle{state}=[
        draw = none,
        fill = white,
        minimum size = 2mm
        ]
        \tikzstyle{fixed}=[
        draw=black,
        rectangle, 
        thick,
        fill = white,
        minimum size = 2mm
        ]

        \node[state] (A) {$A$};
        \node[fixed] (a) [right of=A, xshift=-1cm] {$a$};
        \node[state] (Y) [right of=a] {$Y(a)$};

        \path[->, blue] (a) edge node {} (Y);

    \end{tikzpicture}
    \subcaption{}
    \label{fig:rct-no-cov-2}%
    \end{minipage}%
    \begin{minipage}[l]{0.3\textwidth}
    \centering
    \begin{tikzpicture}[
        > = stealth, 
        auto,
        node distance = 1.5cm, 
        semithick 
        ]

        \tikzstyle{state}=[
        draw = none,
        fill = white,
        minimum size = 2mm
        ]
        \tikzstyle{fixed}=[
        draw=black,
        rectangle, 
        thick,
        fill = white,
        minimum size = 2mm
        ]

        \tikzstyle{hidden}=[
        draw=black,
        circle, 
        thick,
        fill = white,
        minimum size = 2mm
        ]
        \node[state] (A) {$A$};
        \node[state] (Y) [right of=A] {$Y$};
        \node[state] (C) [above of=Y] {$C$};
        \node[hidden] (U) [above of=A] {$U$};

        \path[->, blue] (C) edge node {} (Y);
        \path[->, blue] (U) edge node {} (Y);
        \path[->, blue] (A) edge node {} (Y);
    \end{tikzpicture}
    \subcaption{}
    \label{fig:variables}
    \end{minipage}
    \caption{(a) Causal DAG representation of variables in a randomized experiment where an independent variable $A$ is hypothesized to affect a dependent variable $Y$; (b) The randomized experiment represented as a SWIG, where after the natural value of $A$ occurred, the treatment at a fixed level $a$ was given to the participant instead. The outcome variable under this fixed treatment is $Y(a)$; (c) The randomized experiment with an independent variable $A$, outcome variable $Y$, observed baseline covariates $C$, and unobserved baseline covariates $U$.} 
\end{figure}
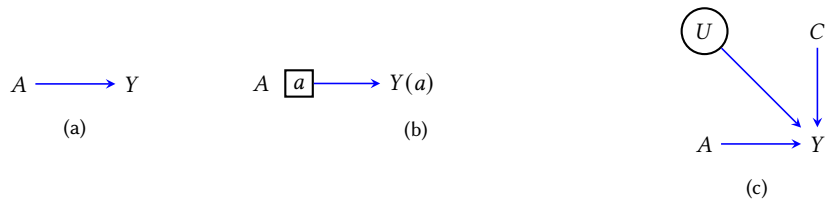
We will use the term \emph{randomized experiment} (as depicted in Fig. \ref{fig:taxonomy}) to mean a \emph{quantitative multi-arm randomized} study conducted in laboratory environments, over a single session.  Most commonly, this corresponds to a between-subjects randomized experiment. We can also think of within-subjects experiments as a randomized experiment, provided that there are no carryover effects from one experimental condition to the next. If substantial carryover effects are suspected, then each subject's experimental conditions should be considered as a longitudinal sequence. This is because carryover effects introduce additional complications for causal inference, as there is a feedback loop between the experimental conditions and the subject's state over time. Careful analysis is needed to avoid mistakes such as those outlined in \cref{sec:existing-methods-fail}. We cover longitudinal problems separately in \cref{sec:longitudinal}

In this section we reintroduce randomized experiments using causal inference language. The purpose of this is to demonstrate how the framework of randomized experiments depends upon (reasonable) assumptions in order for the analyst to derive causal conclusions. In the next section we will detail how those assumptions need to change in order to continue deriving causal conclusions under an observational setting.

The \emph{randomized experiment} is depicted in the causal DAG in Figure \ref{fig:rct-no-cov-1}. For the purposes of exposition, let us consider the example of \cite{gombolayRoboticAssistanceCoordination2016}, in which the authors performed a randomized experiment studying the impact of the type of decision support (computer-issued or robot-issued) on subject trust in the recommendation. We observe two random variables---the type of decision support  encoded as a binary variable $A$ and the subject trust in the recommendation encoded as a rating $Y$. For the moment, let us state that $A$ and $Y$ are the only variables in this example and that we hypothesize that $A$ is a direct cause of $Y$. This immediately implies that $A$ is assigned randomly to participants, since it depends on no variable. In this situation, we normally conclude that because $A$ is randomly assigned, we can simply look at the distribution $p(Y \mid A=a)$ to see what the outcome is under intervention. However, for this section, we will consider the problem in terms of counterfactuals to highlight in detail the steps involved in linking the model to the data to draw causal conclusions. 

In order to check the hypothesis of a causal relationship between $A$ and $Y$ (\ie the presence or absence of the $A \to Y$ arrow), we would like to consider the counterfactual where we intervene on the type of decision support and then observe the effect of this intervention on subject trust. Consider the hypothetical setting where we set the decision support to robot-issued ($A=1$). Then, the response under this set value is called the potential outcome $Y(a=1)$, which is interpreted as the measured subject trust had the decision support been set to robotic, regardless of the actual type of decision support that occurred. 

Note that we do not observe the potential outcomes directly. Indeed, for each participant indexed by $i$, we observe $(A, Y)_{i=1}^N$, where $ Y = Y(a=1)A + (1-A) Y(a=0)$, which means that we can only see one potential outcome at most. This is because we can only assign one of the decision support systems at a time to each participant \footnote{While it is possible to conduct a within-subjects design to assign all conditions in turn, it is not possible to completely eradicate all spillover effects. }In order to move from the observed data $p(Y, A)$ to the counterfactual distribution $p(Y(a))$ we must provide an argument linking the counterfactual to the observed data. 

First, we assume that the measured subject trust when we observe the decision support at a particular level is the same as the measured subject trust had we intervened to set the decision support at that particular level. The intuition behind this assumption is that the act of intervening in the system does not change the way it responds. This assumption is called \emph{consistency}, and it states that
\[Y(a) = Y \textrm{ when $A=a$}\]
Second, we assume that the random independent variable assignment is independent of the potential outcome. That is, the decision support assignment is independent of subject trust in a world where, regardless of the randomly drawn decision support assignment, subject trust had been set to some specified level. This assumption is called \emph{ignorability} and it states that
\[Y(a) \ci A\]
Third, we assume \emph{positivity} of the independent variable assignment. That is, $p(A=a) > 0$ for the desired level of $a$. If $a=1$, this means that we have observed at least one participant get the robot-issued decision support in the randomized experiment.

Given these three assumptions, we can prove that in a randomized experiment, we do in fact obtain the potential outcome  $p(Y(a))$
\begin{align*}
    p(Y(a)) &= p(Y(a) | A=a) \\ 
            &= p(Y | A=a)
\end{align*}
where the first equality holds by the ignorability assumption and the second by the consistency assumption. This returns the intuitive result that in a randomized experiment, computing associational measures (such as the conditional probability $p(Y \mid A=a)$) on the experiment data reveals causal relationships (in this case changing $A$ and observing $Y$).


While we can consider potential outcomes $p(Y(a))$ in isolation, a scientific experiment will more commonly compare the difference between two or more treatment settings. A common target of interest is the difference between the treatment and control (for example, the difference between robot-issued decision support at $A=1$ versus computer-issued decision support at $A=0$). This is represented by $\beta$, and under our assumptions, it is identified as 
\[\beta = \E[Y(a=1)] - \E[Y(a=0)]  = \E[Y \mid A=1] -\E[Y \mid A=0].\]

This quantity $\beta$ is termed the average causal effect, and practitioners will recognize it as the difference in means between the trial arms, which has already been written about in quantitative methods textbooks or review papers such as \cite{hoffmanPrimerConductingExperiments2020}.

\section{Causal Inference for Static Studies} \label{sec:field-studies}
In this section, we turn to non-randomized field studies that are static (see Figure \ref{fig:taxonomy}). We specifically refer to \emph{static} studies rather than cross-sectional studies, as we will consider randomized and non-randomized study designs in both field and laboratory contexts. 
\subsection{Causal Inference with Observed Common Causes between Independent and Dependent Variables} \label{sec:observed-common-causes}
Returning to the scenario from \cite{gombolayRoboticAssistanceCoordination2016}, we consider a variation of the original study concerning the type of decision support system (computer-issued or robot-issued) on subject trust. Suppose that both types of decision support systems were made available in a hospital operating as usual. However, rather than randomly assigning decision support systems, we allowed physicians at the hospital to choose either the existing computer-issued decision support (given by $A=0$) or instead opt for robot-issued decision support (given by $A=1$). Allowing flexibility in physician choice may improve the likelihood of being able to run such a field experiment. However, it is possible that there exists some observed factor $C$---perhaps the physician's baseline trust of decision support systems, which we are able to measure through a short questionnaire administered upon consent to participation. This could bias the physician's choice of decision support towards the computer-issued decision support (the less invasive choice), as well as negatively influence the physician's trust in either of the systems (given by $Y$), regardless of the systems' performance. In this case, it is no longer the case that $p(Y(a)) = p(Y \mid A=a)$, and assuming otherwise can result in bias. We address the potential consequences of failing to account for non-random treatment assignment and describe how it can be corrected under certain assumptions.

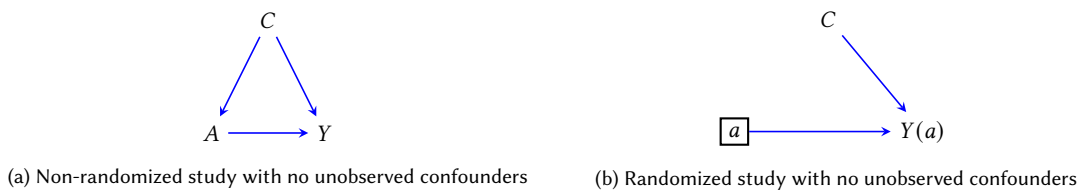
\begin{figure}[h]
    \begin{minipage}[l]{0.5\textwidth}
    \centering
    \begin{tikzpicture}[
        > = stealth, 
        auto,
        node distance = 1.5cm, 
        semithick 
        ]

        \tikzstyle{state}=[
        draw = none,
        fill = white,
        minimum size = 2mm
        ]
        \tikzstyle{fixed}=[
        draw=black,
        rectangle, 
        thick,
        fill = white,
        minimum size = 2mm
        ]

        \node[state] (A) {$A$};
        \node[state] (Y) [right of=A] {$Y$};
        \node[state] (C) [above of=Y, xshift=-.75cm] {$C$};

        \path[->, blue] (A) edge node {} (Y);
        \path[->, blue] (C) edge node {} (Y);
        \path[->, blue] (C) edge node {} (A);
    \end{tikzpicture}
    \subcaption{Non-randomized study with no unobserved confounders}
    \label{fig:obs-study}
    \end{minipage}%
    \begin{minipage}[l]{0.5\textwidth}
        \centering
    \begin{tikzpicture}[
        > = stealth, 
        auto,
        node distance = 1.5cm, 
        semithick 
        ]

        \tikzstyle{state}=[
        draw = none,
        fill = white,
        minimum size = 2mm
        ]
        \tikzstyle{fixed}=[
        draw=black,
        rectangle, 
        thick,
        fill = white,
        minimum size = 2mm
        ]

        \node[fixed] (aa) [left of=a] {$a$};
        \node[state] (Y) [right of=aa, xshift=1cm] {$Y(a)$};
        \node[state] (C) [above of=Y, xshift=-1.25cm] {$C$};

        \path[->, blue] (aa) edge node {} (Y);
        \path[->, blue] (C) edge node {} (Y);
    \end{tikzpicture}
    \subcaption{Randomized study with no unobserved confounders}
    \label{fig:obs-study-swig}
\end{minipage}

\caption{Experimental studies}
\end{figure}

Figure \ref{fig:obs-study} represents the hypothetical study previously described. $C$ is known as a confounder since it induces a spurious correlation between $A$ and $Y$ that is not causal. Figure \ref{fig:obs-study-swig} represents the study we would have liked to perform instead had randomization been possible in the hospital. In this situation, the experimenters assign the decision support type $A=a$, overriding physician preferences. 

To perform this identification linking the observed data to the counterfactual, we make three assumptions. The first is the assumption of \emph{consistency}, where we assume that the counterfactual outcome $Y(a)$ is equal to the observed $Y$ if $A=a$. Second, we assume \emph{conditional ignorability}, which states that $Y(a) \ci A \mid C$---that is, conditioned levels of the observed confounder $C$, the treatment $A$, and the potential outcome $Y(a)$ are not associated. Third, we assume \emph{positivity} at each level of the baseline covariate, which states that $p(A =a \mid C) > 0$ for each level of $a$. Using these three assumptions, it then follows that if we wished to know the causal effect of the decision support system on trust, we can use:

\begin{align}
    p(Y(a)) &= \sum_C p(Y(a) \mid C ) p(C) \label{eqn:application_prob_laws}\\
            &= \sum_C p(Y(a) \mid C, A=a) p(C) \label{eqn:application_cond_ignorability}\\
            &= \sum_C p(Y \mid C, A=a) p(C) \label{eqn:application_consistency}
\end{align}
where the first equality in Eqn. \ref{eqn:application_prob_laws} holds by properties of probabilities, the second in Eqn. \ref{eqn:application_cond_ignorability} holds by conditional ignorability and positivity, and the third in Eqn. \ref{eqn:application_consistency} by consistency.

To summarize, we wanted to learn if computer-issued or robot-issued decision support had an impact on physician trust. For ethical reasons, an experiment using randomization could not be performed. Instead, we collected observational data from a situation where physicians were able to make a choice on the decision support system they used (and we were able to predict that choice given some characteristics such as baseline trust of decision support systems). The counterfactual outcome $Y(a)$ represents physician trust, had we possibly contrary to fact performed an experiment randomly assigning support system types to physicians. This outcome was not measured directly, but rather we used causal inference assumptions to link this to the observational data. One way to explain the intuition behind this method is to notice that by algebraic manipulation,
\begin{align}
    p(Y(a)) &= \sum_C p(Y \mid C, A=a) p(C) \label{eqn:g-formula}\\
            &= \sum_{C} \frac{p(Y, A=a, C) }{p(A=a \mid C)}. \label{eqn:ipw}
\end{align}
The observed data comes from $p(Y, A, C)$, where decision support was not randomly assigned. However, in the counterfactual world, it is randomly assigned. 
Eqns. \ref{eqn:g-formula} and \ref{eqn:ipw} motivate some simple estimation strategies. Eqn. \ref{eqn:g-formula} leads directly to the \emph{g-formula estimator}:
\begin{equation}\hat{\E}_g [Y(a)]  = \frac{1}{n} \sum_{i=1}^n \hat{\E}[Y \mid A=a, C_i] \label{eqn:g-formula-est}
\end{equation}
where $i=1, \ldots, n$ is an index for the samples in the study, $\hat{\E}[Y \mid A, C]$ is a fitted regression model (\eg linear regression, random forests), and $\hat{\E}[Y \mid A=a, C_i]$ represents a prediction of the fitted model for each row $i$ of the data. Eqn. \ref{eqn:ipw} leads to the \emph{inverse propensity-score weighting} (IPW) estimator, which is
\begin{equation} \hat{\E}_{ipw} [Y(a)] = \frac{1}{n} \sum_{i=1}^n \frac{Y_i I(A_i = a) }{\hat{p}(A = a \mid C_i)} \label{eqn:ipw-est}
\end{equation}
where $\hat{p}(A \mid C_i)$ is a model (\eg a logistic regression) for the propensity score evaluated at the covariates of the $i$th sample, $C_i$.

Both of these estimators lead to unbiased estimation of the counterfactual outcome under intervention $\E[Y(a)]$. Furthermore, each estimator offers a different interpretation of how to move from the first world to the second. Eqn. \ref{eqn:g-formula-est} says that we must look at the average outcome given each strata of decision support assignment and physician baseline characteristics. Then, if we take a weighted average of the strata by the frequency of the physician characteristics, we obtain the causal effect of that decision support assignment. Eqn. \ref{eqn:ipw-est} says that we must reweigh the observed data by the probability of the physician's choice given the observed physician characteristics $p(A \mid C)$. Intuitively, this ``cancels out'' the non-randomized decision support assignment and gives us a distribution from the counterfactual world. Using this idea of controlling for confounding, we investigate a hypothetical HRI scenario motivated by a real research problem.
\subsubsection{Further reading: Causal identification theory}
There has been considerable research into the theory of linking observed data distributions to the counterfactual. An early result that generalizes \cref{eqn:g-formula} appeared as the \emph{backdoor criterion} \cite{pearlBayesianAnalysisExpert1993}, which is a sound algorithm for establishing causal effects.

\begin{dfn}(Backdoor criterion) \cite{pearlBayesianAnalysisExpert1993,pearlCausality2009} \label{dfn:backdoor}
A set of variables $\mathbf{W}$ satisfies the backdoor criterion relative to intervention $A$ and outcome $Y$ if no node in $\mathbf{W}$ is a descendant of $A$, and $\mathbf{W}$ blocks every path between $A$ and $Y$ that contains an arrow into $A$.

If this criterion is satisfied, then $p(Y(a))$ is identified via
\[p(Y(a)) = \sum_{\mathbf{W}} p(Y \mid A=a, \mathbf{W} ) p(\mathbf{W}).\]
\end{dfn}

A general algorithm for the identification of causal effects has been discovered and proved complete (see \cite{shpitserIdentificationJointInterventional2006,huangPearlCalculusIntervention2006}).

\subsubsection{Case study: Social robotics in the field}
The work presented in \cite{moshkinaSocialEngagementPublic2014} involves an observational field study in which a robot explores a public venue. The primary hypothesis was that machines that present consistent human-like characteristics are more likely to invoke a social response. The study was conducted using a between-subjects design, where the independent variables were the type of short story recited by the robot (\emph{humorous,  informative}), and levels of social cues (\emph{no movement} to \emph{full body movement}). The outcome of interest was attention retention, which was measured by the number of people that observed the robot for at least 15 seconds.

A key aspect of the study was that the experimental setup was positioned next to two modern US Navy ships and several other exhibits. Therefore, the experimental setup had to compete against the other exhibits to attract participants, and the activity levels at other exhibits were likely to influence the outcome variable of attention retention. To see how causal inference may be applicable, we introduce a hypothetical but plausible twist to the experimental setup. Imagine that the robot was able to detect the activity levels around its setup. Then, the robot could respond to reduced activity by engaging in more exaggerated social behaviors such as larger body movements. Under this setup, if we wanted to learn which social cues were more effective at attention retention, we would not be able to deduce this directly from the data.

We can consider this setup using the graph in Figure \ref{fig:obs-study}, where $A$ denotes the social behaviors of the robot, $C$ denotes the activity levels at other exhibits, and $Y$ denotes the attention retention of the audience. The $A \to Y$ edge exists because the social behaviors of the robot affect the attention retention of the audience. Furthermore, the $C \to Y$ edge exists because we posit that the activity levels at other exhibits influence the attention retention of the audience, as they may be more distracted under busier conditions. Finally, the $C \to A$ edge exists because of the responsive nature of the robot to its environment. In this scenario, we can learn which levels of social cues are more effective at retaining attention overall by estimating $\E[Y(a)]$ using the techniques outlined earlier 
in Section \ref{sec:observed-common-causes}.

\subsection{Causal Inference for Non-Randomized Studies with Unobserved Variables}\label{sec:unobserved_confounders}
So far, we have described contexts where all causally relevant variables in the problem were assumed to be observable and measurable. In practice, most practitioners would acknowledge this assumption to be false, since it is usually quite difficult to rule out all possible unobserved variables a priori. Depending on the purpose of the analysis, unobserved variables can be relatively benign. For example, many machine learning algorithms might posit latent variables during inference. These latent variables are unobserved but do not otherwise interfere in learning associational relationships between the observed variables.

However, unobserved variables are potentially catastrophic in causal inference. This is because unobserved variables can alter the causal conclusions of a problem in such a way as to be invisible to the practitioner who only has knowledge of the observed variables. Furthermore, attempts to use machine learning or other tools to try to circumvent this limitation have no guarantees of success unless further assumptions are made about the problem (\eg observing more variables, introducing more structure to the model). 

\begin{figure}[h]
    \centering
    \begin{tikzpicture}[
        > = stealth, 
        auto,
        node distance = 1.5cm, 
        semithick 
        ]

        \tikzstyle{state}=[
        draw = none,
        fill = white,
        minimum size = 2mm
        ]
        \tikzstyle{fixed}=[
        draw=black,
        rectangle, 
        thick,
        fill = white,
        minimum size = 2mm
        ]

        \tikzstyle{hidden}=[
        draw=black,
        circle, 
        thick,
        fill = white,
        minimum size = 2mm
        ]
        \node[state] (A) {$A$};
        \node[state] (Y) [right of=A] {$Y$};
        \node[hidden] (C) [above of=Y, xshift=-.75cm] {$U$};

        \path[->, blue] (A) edge node {} (Y);
        \path[->, blue] (C) edge node {} (Y);
        \path[->, blue] (C) edge node {} (A);
    \end{tikzpicture}
    \caption{Bow-arc problem modeling observational study with unobserved confounders. In the context of the example from \cite{gombolayRoboticAssistanceCoordination2016}, $A$ denotes the type of decision support system chosen by physicians, $Y$ denotes physician trust in the chosen system, and $U$ denotes some unobserved factor that affects both physician trust and choice of decision support system.}
    \label{fig:study-unobs-confounders}
\end{figure}
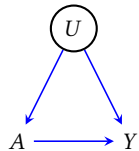

Non-randomized studies involving unobserved variables can be modeled as the \emph{bow-arc problem} from causal literature, so-named because there is the frame of the bow (through an unobserved confounder $U$), and the string of the bow (through the direct edge) (Figure \ref{fig:study-unobs-confounders}). 

Returning to the previous example of \cite{gombolayRoboticAssistanceCoordination2016}, we posit that, in addition to the existing variables $A$ (denoting the type of decision support system) and $Y$ (physician trust in the chosen system), there was some unobserved variable $U$ that affected both $A$ and $Y$. For example, $U$ could be the physician's baseline trust of decision support systems, but unlike the example in section \ref{sec:observed-common-causes}, no survey was taken of the physician's baseline trust. $U$ could also be the physician's mood---perhaps they are less patient that day and unwilling to try a new decision support system and less willing to put effort into building a trusting relationship with support systems generally. 

Absent further assumptions, an observational study collecting only $A$ and $Y$ will not be able to identify the causal effect of $A$ on $Y$. The intuitive explanation for this is that there are two ways $A$ can affect $Y$---through the path $A \leftarrow U \rightarrow Y$ or through the path $A \to Y$. However, the observed data contains only $A$ and $Y$. This does not let us distinguish the confounding of $U$ from the direct effect of $A$ on $Y$, since in truth, it may be the case that there is a strong correlation due to $U$ and no direct effect or no correlation and a strong direct effect, both of which could generate the observed data.

The implication of this result is that if we are unable to rule out plausible unobserved confounders between variables of interest, it may impede causal inference, and we may not be able to provide an estimator for the causal effect. \cite{shpitserIdentificationJointInterventional2006} provides a more formal proof of the claim made above and an algorithm that provides the precise conditions under which unobserved confounders will impede causal inference in a particular causal graph. If such limitations are encountered, an analyst can either attempt to collect more data (to alter the graph) or ask a different question (to change the causal effect).


\subsection{Transporting Inferences between Different Domains}
\label{sec:transportability}
Scientific results that hold in one context (cultural, societal, economic, or otherwise) do not always generalize to other contexts. This idea has been discussed in terms of transportability, meta-analysis, external validity, and quasi-experiments \cite{pearlTransportabilityCausalStatistical2011}. For example, results that are obtained from performing studies in a population of people from one cultural background might not generalize immediately to other populations. Cultural background can affect people's perceptions of robots and robot behavior \citep{limSocialRobotsGlobal2020}. 

The work described in \cite{wangWhenRomeRole2010} studied the extent to which university students from China and the United States would respond to a robot's recommendation. A key finding from the study was that Chinese participants were less receptive to robot recommendations compared to American participants. This suggests that if we were to perform an observational study on two competing technologies in China, we might not be able to immediately generalize these results to an American context. This example highlights that cultural contexts are an impact factor to consider in causal inference.

Building off this example, consider a hypothetical scenario where we can conduct a randomized experiment in China to understand the causal effect of two competing decision support technologies. In this scenario, we want to compute the causal effect in the United States, and we are limited to collecting baseline covariate information in the United States. 
The causal graph is identical to that of Figure \ref{fig:obs-study}. Let $A$ denote the type of decision chosen by the participant (computer-issued or robot-issued), $C$ denote the receptiveness of the participant to robot-issued support, and $Y$ denote the participant's rating of the decision support they used. The distinction in this case is that we have two different distributions---$p(\cdot)$ denotes data from China and $p^*(\cdot)$ denotes data from the United States. Assume that we have experimental information at different rating levels $p(Y \mid A, C)$ from an experiment performed in China and that the mechanism of rating is the same within strata of receptivity to robot-issued support in the United States and China, which means that $p(Y \mid A, C ) = p^*(Y \mid A, C)$. Furthermore, we assume that we have access to the distribution of receptivity to robot-issued decision support $p^*(C)$ through survey data from the United States. Then, we can compute the population causal effect of the decision support system on participant rating in the United States, $p^* (Y(a))$:

\begin{align*}
p^*(Y(a)) &= \sum_{C} p^*(Y(a) \mid C) p^*(C)\\
&= \sum_{C} p^*(Y \mid A=a, C)p^*(C) \\
&= \sum_{C} p(Y \mid A=a, C)p^*(C) \\
\end{align*}
where the first equality follows by the laws of probability; the second by positivity, conditional ignorability, and consistency; and the third by the shared responsiveness within strata of receptivity in the United States and China. Intuitively, this result is similar to Eqn. \ref{eqn:g-formula}. Since we wanted to compute a result in the United States, we used distributions $p^*(\cdot)$ that were from the United States. However, because we didn't have access to an experiment or a full observational study, we instead borrowed data from China to obtain valid inference in the United States.

\subsection{Measurement Error}\label{sec:measurement_bias}
Measurement error occurs when a phenomena of interest cannot be measured directly, but a proxy of that phenomena can be measured, albeit imperfectly. If this imperfection is sufficiently small to be ignored, we can proceed by assuming that the proxy and the phenomena of interest are one and the same. However, sometimes the measurement error of the proxy is too large to ignore. This can result in biased estimates in causal inference and in inference in general.

HRI studies can include variables that are important to investigate but difficult to measure. For instance, variables such as trust are difficult to measure objectively, although psychophysiological methods of determining trust are being developed \cite{ajenaghughrureMeasuringTrustPsychophysiological2020}. In practice, scales based on participant surveys have been devised to provide an approximate measure of these variables \cite{gulatiDesignDevelopmentEvaluation2019}. However, these surveys are not error-free measurements of trust. 

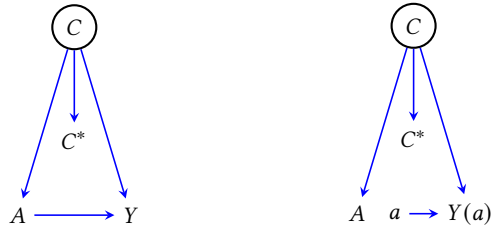
\begin{figure}[t]
\begin{minipage}{.3\textwidth}
    \centering
    \begin{tikzpicture}[
        > = stealth, 
        auto,
        node distance = 1.5cm, 
        semithick 
        ]

        \tikzstyle{state}=[
        draw = none,
        fill = white,
        minimum size = 2mm
        ]
        \tikzstyle{fixed}=[
        draw=black,
        rectangle, 
        thick,
        fill = white,
        minimum size = 2mm
        ]

        \tikzstyle{hidden}=[
        draw=black,
        circle, 
        thick,
        fill = white,
        minimum size = 2mm
        ]
        \node[state] (A) {$A$};
        \node[state] (Y) [right of=A] {$Y$};
        \node[hidden] (C) [above of=Y, yshift=1cm, xshift=-.75cm] {$C$};
        \node[state] (Cp) [below of=C] {$C^*$};

        \path[->, blue] (A) edge node {} (Y);
        \path[->, blue] (C) edge node {} (Y);
        \path[->, blue] (C) edge node {} (A);
        \path[->, blue] (C) edge node {} (Cp);
    \end{tikzpicture}
    
\end{minipage}
\begin{minipage}{.3\textwidth}
    \centering
    \begin{tikzpicture}[
        > = stealth, 
        auto,
        node distance = 1.5cm, 
        semithick 
        ]

        \tikzstyle{state}=[
        draw = none,
        fill = white,
        minimum size = 2mm
        ]
        \tikzstyle{fixed}=[
        draw=black,
        rectangle, 
        thick,
        fill = white,
        minimum size = 2mm
        ]

        \tikzstyle{hidden}=[
        draw=black,
        circle, 
        thick,
        fill = white,
        minimum size = 2mm
        ]
        \node[state] (A) {$A$};
        \node[state] (AA) [right of=A, xshift=-1cm] {$a$};
        \node[state] (Y) [right of=A] {$Y(a)$};
        \node[hidden] (C) [above of=Y, yshift=1cm, xshift=-.75cm] {$C$};
        \node[state] (Cp) [below of=C] {$C^*$};

        \path[->, blue] (AA) edge node {} (Y);
        \path[->, blue] (C) edge node {} (Y);
        \path[->, blue] (C) edge node {} (A);
        \path[->, blue] (C) edge node {} (Cp);
    \end{tikzpicture}
    
\end{minipage}
    \caption{Measurement error scenario. Let $A$ be some experimental conditions and $Y$ be some outcome metric. In this observational study setting, we assume that the true but unobserved user trust $C$ influences both the treatment choice and the outcome. However, we can measure $C^*$, which represents a scale devised to measure trust. The graph on the left denotes the observed data distribution, while the graph on the right denotes the single-world intervention graph (SWIG) with the desired counterfactual $Y(a)$ denoting the outcome metric, had the experimental conditions been set such that $A = a$.}
    \label{fig:measurement-bias}
\end{figure}

To see why this is an issue, consider Figure \ref{fig:measurement-bias}, where $C$ represents the true latent trust and $C^*$ represents a scale constructed to measure trust. Clearly, the latent trust influences the proxy constructed to measure it, which justifies the edge $C \to C^*$. By the rules of d-separation, we know that conditioning on $C^*$ does not block the flow of association through the path $A \leftarrow C \rightarrow Y$ if $C^*$ is not equal to $C$. Thus, conditioning on $C^*$ does not provide for identification. 

To perform causal inference given proxies, we first need some estimate of the measurement error associated with $C^*$ and $C$. In the simple case where both of these variables are binary, this reduces to the problem of estimating misclassification error rates. For example, this could be done by running an experiment to determine the accuracy of the trust scale compared to the psychophysiological measurement of trust, $p(C^* \mid C)$. Then, under a non-differential error assumption (\ie the latent trust is the only factor that influences the measured trust), we obtain:

\begin{align*}
    p(Y, A, C^* ) &= \sum_{C} p(Y, A, C, C^*) \\
                 &= \sum_{C} p(C^* \mid Y, A, C) p(Y, A, C) \\
                 &= \sum_{C} p(C^* \mid C) p(Y, A, C)
\end{align*}

Viewing this as a linear algebra problem, we note that $p(C^* \mid C)$ can be viewed as a stochastic matrix of full rank, provided that the cardinality of $C^*$ is greater than or equal to that of $C$, that therefore has an inverse denoted $M(C^*, C)$. Thus,

\begin{align} \label{eqn:measurement-bias-invert}
    p(Y,A,C) &= \sum_{C^*} M(C^*, C) p(Y, A, C^*) 
\end{align}
Substituting Eqn. \ref{eqn:measurement-bias-invert} into Eqn. \ref{eqn:ipw} from Section \ref{sec:observed-common-causes} will result in the following equation:
\begin{align}
    p(Y(a)) =\sum_{C} \sum_{C^*} M(C^*, C) p(Y, A, C^*) \frac{\sum_C M(C^*, C) p(C)}{\sum_C M(C^*, C) p(C, A)} \label{eqn:measurement-bias}
\end{align}
This result means that we are able to perform valid causal inference even if we only have a noisy proxy of the true confounder $C^*$, so long as we have some measure of its misclassification rate $p(C^* \mid C)$. Noisy proxies are often required to measure social or cultural phenomena in HRI studies, so handling them correctly will be important in field or observational studies.

\subsection{Causal Discovery}
\label{sec:causal-discovery}
Thus far, we have described a series of methods for performing inferences using graphs. Constructing these graphs is often done using domain knowledge to argue for the absence or presence of edges, but in certain scenarios, we may not know enough about a particular domain to do so. \emph{Causal discovery} methods allow for data-driven approaches to drawing a causal graph. These methods exploit information about conditional independence in the data to decide on the absence or presence of edges. However, in many cases, the information contained in the data is not sufficient to recover a DAG because the orientation of edges is not always possible---we may know that an edge exists, but figuring out which way it points is not always possible.

To illustrate this difficulty, we revisit Figure \ref{fig:d-sep} from Section \ref{sec:graphical-models}---it turns out that given data generated from any of these graphs, we can only tell whether the data is from the collider graph or not. This is because, for both the chain and fork graphs, the only conditional independence statement that holds is $X \ci Y \mid Z$, whereas for the collider graph, the only conditional independence statement that holds is $X \ci Y$. Thus, the three DAGs form two equivalence classes, one containing only the collider graph, and the other containing both the chain and the fork graphs. An algorithm can empirically validate if these conditional independence statements hold in the data and select the appropriate equivalence class. Note that the equivalence classes can contain more than one DAG, and thus more than one causal explanation, for a particular dataset.

The PC \cite{spirtesCausationPredictionSearch2000} and GES \cite{chickeringOptimalStructureIdentification2002} algorithms recover equivalence classes of DAGs. These algorithms assume that there are no unobserved confounders (\ie all causally relevant variables are observed in the data). If unobserved confounders are suspected to be present, then the FCI or FGES algorithms should be used instead (see \cite{shenChallengesOpportunitiesCausal2020a} for an example using data on Alzheimer's disease).

\section{Causal Inference for Longitudinal Studies} \label{sec:longitudinal}

Longitudinal studies are useful for investigating many HRI questions. The implications of many proposed HRI technologies are only fully exhibited in a longer-term study in which the interaction loses its novelty. The need for longitudinal research is already well-recognized by the HRI community \cite{fischerEffectConfirmedPatient2021}. In this section, we outline how causal inference can analyze data from the \emph{field}, in which experimental conditions may not be randomly assigned, and where repeated actions influence the subject's state over multiple time periods. 

\subsection{Current State of Longitudinal HRI Research}
We consider examples of longitudinal HRI studies. The work described in \cite{kandaTwoMonthFieldTrial2007} involves a two-month field trial of a social robot deployed in a primary school. The social robot is equipped with the ability to form long-term relationships and estimate friendly relationships among people. By recognizing participants as unique entities, it is able to adapt its behaviors to each participant and modify its behaviors over repeated interactions. The robot was able to establish friendly relationships with many of the children over two months. Other examples of longitudinal HRI research can be seen in \cite{bohusManagingHumanRobotEngagement2014}, which describes an in-the-wild study in which a direction-giving robot is deployed over a period of five days, and \cite{gordonAffectivePersonalizationSocial2015}, which evaluates an integrated second language-learning game and autonomous social robotic learning companion 
in the wild with 34 children over two months.

The experiments in these examples are characterized as exploratory and not hypothesis-driven. In each example, a new technology is developed and deployed in a real-world context to better understand how human-robot interactions occur in the wild. However, the new technology is not compared to anything else, and the outcomes are usually qualitative in nature.  It is difficult to draw insights from such data that can immediately drive changes in policy or practice. Furthermore, longitudinal HRI research presents various challenges. Longitudinal studies are about particular phenomena that occur only in the field and are difficult to replicate in a controlled laboratory environment. They often take weeks or months to perform, which makes them relatively costly exercises. Furthermore, robots in the field require more sophistication to operate safely and correctly, which is not always possible with the current state of technology. In the future, we expect that the state of the art in robotics will advance to the point where robots exist in the wild (\eg household robots such as Amazon Astro). This may lead to vast quantities of data collected over time in non-laboratory settings. Applying causal inference techniques to such data could result in powerful new insights that are difficult to obtain with traditional experiments.

\subsection{Why Existing Methods Fail for Time-Varying Interventions}\label{sec:existing-methods-fail}

Longitudinal HRI research involves a robot and human interacting repeatedly. Both the robot and the human can change their behavior based on what the other is doing or has done in the past. This temporal dependency can result in incorrect conclusions if causal methods are not used. As an example, we recast the work described in \cite{hernanCausalInferenceWhat2020}  in terms of social navigation, based on the work in \cite{mavrogiannisEffectsDistinctRobot2019}. In this example, a robot is attempting to navigate through a heavily trafficked area and has two behavior modes that it can choose from, captured by $A_0 \in \{0, 1\}$. Based on the feedback and social cues it detects from nearby people (captured as $L_1$), the robot can choose to change its activity (captured as $A_1$). A rating of the human participant's final impressions is recorded as $Y$.

\begin{figure}[t]
    \centering
    \begin{tikzpicture}[
        > = stealth, 
        auto,
        node distance = 1.5cm, 
        semithick 
        ]

        \tikzstyle{state}=[
        draw = none,
        fill = white,
        minimum size = 2mm
        ]
        \tikzstyle{fixed}=[
        draw=black,
        rectangle, 
        thick,
        fill = white,
        minimum size = 2mm
        ]
        \tikzstyle{hidden}=[
        draw=black,
        circle, 
        thick,
        fill = white,
        minimum size = 2mm
        ]

        \node[state] (A0) {$A_0$};
        \node[state] (L1) [right of=A0] {$L_1$};
        \node[state] (A1) [right of=L1] {$A_1$};
        \node[state] (Y) [right of=A1] {$Y$};
        \node[hidden] (U) [below of=A1] {$U$};

        \path[->, blue] (A0) edge node {} (L1);
        \path[->, blue] (L1) edge node {} (A1);
        \path[->, blue] (U) edge node {} (L1);
        \path[->, blue] (U) edge node {} (Y);
    \end{tikzpicture}
    \caption{Example of a longitudinal two-phase study with $A_0$ representing the robot's initial behavior, $L_1$ representing pedestrians' feedback and social cues, which then determines the robot's subsequent behavior $A_1$, and an unobserved variable $U$ that influences $L_1$ and $Y$, which represents the pedestrians' final impressions.}
    \label{fig:longitudinal_fail}

\end{figure}
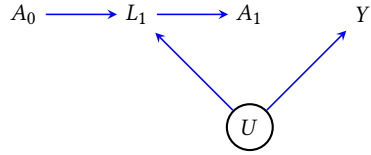

To demonstrate how associational methods can fail, we assume a particular scenario. In this scenario, let $U$ denote a participant's deeply and closely held beliefs about robots. This belief can influence the participant's feedback cues $L_1$ and final impressions $Y$. However, it happens that the participants have such strong beliefs $U$ that this is the only factor that affects their final impression of the robot regardless of how the robot performed. This scenario is depicted in Figure \ref{fig:longitudinal_fail}, where the final impression $Y$ has no incoming edges except for beliefs $U$ and the robot's behaviors $A_0$ and $A_1$ do not influence $Y$.

If we want to discover the optimal robot behavior, we may compare two simple strategies: always using the first type of behavior ($A_0 = 0, A_1=0)$ or always using the second type of behavior ($A_0 = 1, A_1 = 1$). We would then seek to measure the effect $\beta = \E[Y(a_0=1, a_1=1)] -  \E[Y(a_0=0, a_1=0)]$. In truth, in the scenario descibed above, there is no directed edge or directed path (\ie a sequence of directed edges that all point the same way) from either $A_0$ or $A_1$ to $Y$. Therefore, we know that, given this model, a hypothetical experiment where $A_0$ and $A_1$ are manipulated would give a causal effect of $\beta = 0$. However, if a naive analyst fitted a regression $\hat{\E}[Y \mid A_0, A_1]$ (\eg a suitably flexible regression such as a linear regression, random forest, or some other machine learning method) and evaluated $\tilde{\beta} = \hat{\E} [Y \mid A_0 = 1, A_1 = 1] - \hat{\E}[Y \mid A_0 = 0, A_1 = 0] $, they are not guaranteed that $\tilde{\beta} = 0$ even if in truth $\beta = 0$, which can be seen through application of d-separation: First, note that $L_1$ is a collider for the path $A_0 \rightarrow L_1 \leftarrow U \rightarrow Y$. By the rules of d-separation, conditioning on a descendant of a collider (\eg $A_1$) induces dependence between variables (\eg $A_0$ and $Y$). This means there exists a spurious correlation between $A_0$ and $Y$ even though there exists no causal mechanism or pathway from $A_0$ to $Y$. The key takeaway from this example is that even simple two-phase longitudinal problems can be analyzed incorrectly when applying standard analytical techniques such as regression. In the next section, we consider methods that can enable correct analysis of time-varying interventions.


\subsection{Inferring Policy Values from Data}
\label{sec:policy}
In this section, we consider methods for inferring outcomes in longitudinal settings where we can observe but not intervene. Section \ref{sec:existing-methods-fail} outlined how standard methods applied in such settings could fail to provide correct causal conclusions. In this section, we outline how to correctly perform causal inference for longitudinal scenarios. We will consider a two-phase study to motivate the discussion, although the principles discussed here extend to arbitrary numbers of phases. 

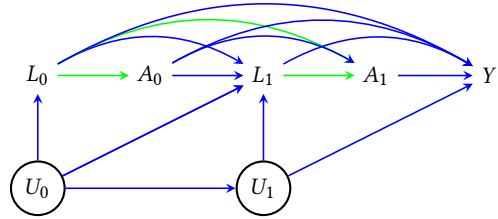
\begin{figure}[t]
    \centering
    \begin{tikzpicture}[
        > = stealth, 
        auto,
        node distance = 1.5cm, 
        semithick 
        ]

        \tikzstyle{state}=[
        draw = none,
        fill = white,
        minimum size = 2mm
        ]
        \tikzstyle{fixed}=[
        draw=black,
        rectangle, 
        thick,
        fill = white,
        minimum size = 2mm
        ]
        \tikzstyle{hidden}=[
        draw=black,
        circle, 
        thick,
        fill = white,
        minimum size = 2mm
        ]

        \node[state] (A0) {$A_0$};
        \node[state] (L0) [left of=A0] {$L_0$};
        \node[state] (L1) [right of=A0] {$L_1$};
        \node[state] (A1) [right of=L1] {$A_1$};
        \node[state] (Y) [right of=A1] {$Y$};
        \node[hidden] (U1) [below of=L1] {$U_1$};
        \node[hidden] (U0) [below of=L0] {$U_0$};

        \path[->, blue] (A0) edge node {} (L1);
        \path[->, green] (L1) edge node {} (A1);
        \path[->, blue] (U1) edge node {} (L1);
        \path[->, blue] (U1) edge node {} (Y);
        \path[->, blue] (A1) edge node {} (Y);
        \path[->, green] (L0) edge node {} (A0);
        \path[->, blue] (U0) edge node {} (L0);
        \path[->, blue] (U0) edge node {} (L1);
        \path[->, blue] (U0) edge node {} (L1);
        \path[->, blue] (U0) edge node {} (U1);
        
        \path[->, blue, bend left=30] (L0) edge node {} (L1);
        \path[->, green, bend left=30] (L0) edge node {} (A1);
        \path[->, blue, bend left=30] (L0) edge node {} (Y);
        \path[->, blue, bend left=30] (A0) edge node {} (A1);
        \path[->, blue, bend left=30] (A0) edge node {} (Y);
        \path[->, blue, bend left=30] (L1) edge node {} (Y);
    \end{tikzpicture}
    \caption{Longitudinal two-phase observational study. The blue edges denote all mechanisms involving human participants. The green edges denote mechanisms that involve the robot's decision making capabilities.}
    \label{fig:longitudinal_case_study}

\end{figure}

Consider a social navigation scenario similar to the one described in Section \ref{sec:existing-methods-fail}. In this example, suppose a company manufactures and sells a line of household robots and that their customers have consented to allow data collection from these robots. The company seeks to improve the social navigation capabilities of their robots using the collected data. In particular, they want to compare two different navigation strategies to find which is preferred by the customers. Let  $L_0$ be some baseline observed covariates, such as the initial behaviors of the human participants, and $U_0$ be some baseline unobserved covariates, such as undisclosed participant beliefs about robots or navigation preferences. The variable $A_0$ represents the first navigation strategy taken by the robot in response to $L_0$ and possibly $U_0$. This action influences the participant's intermediate response (\eg nonverbal feedback, social cues), which is recorded as $L_1$, and may cause the participant to update their beliefs or preferences in the unobserved covariates $U_1$. The variable $A_1$ represents the second navigation strategy taken by the robot, which is made in response to $U_1$ and $L_1$. Finally, a study outcome of the participant's impressions $Y$ are measured, either through additional detection of participant nonverbal or verbal feedback or through  complaints or compliments the participant submitted regarding the robot behavior. This scenario is depicted in Figure \ref{fig:longitudinal_case_study}.


Suppose that the company is interested in learning the outcome of a hypothetical experiment where the robot's navigation strategies are set arbitrarily at each phase. This is represented by the potential outcome $\E[Y(a_0, a_1)]$, the expected value of the outcome variable had $A_0$ been set to value $a_0$ and $A_1$ been set to value $a_1$. It turns out that by extending the g-formula provided in Eqn. \ref{eqn:g-formula} from Section \ref{sec:observed-common-causes}, we obtain:

\begin{align}
\E[Y(a_0, a_1)] &= \sum_{l_0, l_1} \E[Y \mid A_0=a_0, A_1 = a_1, L_0 = l_0, L_1=l_1] p(L_0=l_0) p(L_1=l_1 \mid A_0=a_0, L_0=l_0) \label{eqn:longitudinal-g-formula-2-steps}
\end{align}
The assumptions involved in Eqn. \ref{eqn:longitudinal-g-formula-2-steps} include those encoded within the graph of Figure \ref{fig:longitudinal_case_study}, as well as positivity (\eg we observe all levels of the decisions $A_0, A_1$ and observed covariates $L_0, L_1$). 
To interpret this result, note that this equation first contains a regression model $\E[Y \mid A_0, A_1, L_0, L_1]$ evaluated at actions $A_0 = a_0$ and $A_1 = a_1$. The regression model is averaged over the levels of the participant feedback $L_0$ and $L_1$, weighted by the probabilities $p(L_0) p(L_1 \mid L_0, A_0=a_0)$. These weights can be viewed as the proportions at which each level of $L_0$ and $L_1$ occurs when every participant experiences the same robot behavior in $A_0=a_0, A_1 = a_1$. To estimate $\E[Y(a_0, a_1)]$, we first estimate a regression model for $\E[Y \mid A_0, A_1, L_0, L_1]$. We can estimate $p(L_0)$ by computing an empirical distribution and $p(L_1 \mid A_0, L_0)$ by using a model (if $L_1$ is binary, then a logistic regression model may be used, for instance). The regression models could be linear regressions, logistic regressions, random forests, or other unbiased methods for estimating a conditional expectation, depending on the type of variables involved (continuous or discrete). The equation for the corresponding estimator is
\begin{align}
\hat{\E}[Y(a_0, a_1)] &= \sum_{l_0, l_1} \hat{\E}[Y \mid A_0=a_0, A_1 = a_1, L_0 = l_0, L_1=l_1] \hat{p}(L_0=l_0) \hat{p}(L_1=l_1 \mid A_0=a_0, L_0=l_0), \label{eqn:longitudinal-g-formula-2-steps-estimator}
\end{align}
where each regression or distribution has been replaced by a suitable fitted model. This equation will allow us to compare the difference between a static robot behavior policy where the first navigational strategy is always used ($\hat{E}[Y(a_0=0, a_1 = 0)]$) and a policy where the second navigational strategy is always used ($\hat{E}[Y(a_0=1, a_1 = 1)]$).

In practice, robot behaviors may not be static and can take into account past observations to choose the next action. We introduce policy functions
 $g_0 = g_0(A_0 \mid L_0), g_1= g_1(A_1 \mid A_0, L_0, L_1)$, where each function denotes the conditional probability of taking an action given all information preceding the decision. In other words, the $g$ function represents the probability that the robot takes a particular action having observed a certain pattern of behavior. Incorporating the policies into the expected value of the outcome variable gives us: 

\begin{align}
    \E[Y(g_0, g_1))] &= \sum_{a_0, a_1, l_0, l_1} \E[Y \mid A_0=a_0, A_1 = a_1, L_0 = l_0, L_1=l_1] p(L_0=l_0) p(L_1=l_1 \mid A_0=a_0, L_0=l_0)\nonumber \\
                       & \quad \times g_0 (A_0=a0 \mid L_0=l_0) g_1 (A_1=a_1 \mid A_0=a_0, L_0=l_0, L_1=l_1), \label{eqn:longitudinal-policy-2-steps}
\end{align}
which is a weighted average of $\E[Y \mid A_0, A_1, L_0, L_1]$ over levels of $A_0, A_1, L_0, L_1$, where the weights are given by the probability of each level: 
\[p_g(A_0, A_1, L_0, L_1) = p(L_0=l_0) p(L_1=l_1 \mid A_0=a_0, L_0=l_0) g_0 (A_0=a0 \mid L_0=l_0) g_1 (A_1=a_1 \mid A_0=a_0, L_0=l_0, L_1=l_1).\] 
Note that this weight depends on the policy functions $g_0, g_1$, so different robot behaviors may cause different human behaviors and thus may produce different expected rewards $\E[Y(g_0, g_1)]$. For further details, \cite{hernanCausalInferenceWhat2020} provides more details about estimation of causal effects in the general case of an arbitrary number of phases (Part III, Chapters 20--21).

\subsubsection{Case study: Child-robot tutoring}
Child-robot tutoring shows promise in enabling personalized learning. Robot tutoring systems have the potential to engage students in ways that account for unobserved characteristics of a student, such as motivation. An example of an interactive robot tutoring system can be seen in \cite{ramachandranEffectiveRobotChild2019}, which describes a robot tutor that responds to observed behaviors, such as help-aversion or help-overuse, by providing hints to improve overall learning outcomes. This was done using two rules---if a student makes two incorrect attempts, a hint is provided and if a student makes three consecutive hint requests without attempting the problem, the third hint request is denied until an attempt is made. The personalized tutoring system that followed the two rules to actively provide or withhold hints was compared to a control system in which hints were freely available. A user study found that users of the personalized robot tutoring system significantly decreased sub-optimal help-seeking behaviors and improved test scores compared to users of the control system. While the study was conducted longitudinally over four sessions, the two experimental conditions were held unchanged over the duration of the study.

One extension of this study could be to have the robot tutoring system adapt its hint-providing behavior after each session. Given that students are likely to be either help-averse or help-dependent but not both, it would make sense to tailor the thresholds of these interventions based on student performance. For highly motivated students, it may make sense to have light-touch interventions where no hints are provided automatically, but all hints are provided as requested, while less motivated students might require stricter interventions to encourage them to increase their effort. Naturally, we may ask what the most effective intervention policy is, given that we cannot observe motivation directly. We recast this problem as a longitudinal system. Let $A_0$ and $A_1$ be the first and second robot tutoring interventions. Let $U_0$ be the student's initial latent motivation  and $U_1$ be the student's latent motivation after the first tutoring intervention. Let $L_0$ be the baseline test result (where we are interested in measuring the student's help-seeking behaviors), $L_1$ be the test result after intervention $A_0$, and $Y$ be the test result after both interventions.

Depending on the type of study, we can end up with different graphs based off the graph shown in Figure \ref{fig:longitudinal_case_study}. One way we could set up an experiment is to simply consider each set of interventions $(A_0, A_1)$ as a larger intervention in its own right and to randomize accordingly. One limitation of this approach is that it may put students into sub-optimal intervention policies. This can be a problem in larger trials, where it may be unethical to provide benefit to only a small learning population. 
An alternate experiment setup is a sequentially randomized experiment in which the experimental conditions at $A_0$ and $A_1$ each depend on previous observed behaviors and experimental conditions. Under this setup, students who are unresponsive to the initially assigned intervention have a probability of being switched to a different intervention. A third approach is to conduct the study as an observational study where the experimental conditions are left to the students to select, which may be influenced by their motivation.

Regardless of the study setup, the scientific questions we ask remain the same. First, as in \cite{ramachandranEffectiveRobotChild2019}, we could consider measuring the effect of a static strategy of ``intervention'' versus ``no intervention'' on the study outcome, which is represented as $\E[Y(a_0=1, a_1 = 1)] - \E[Y(a_0=0, a_1 = 0)]$ (using Eqn. \ref{eqn:longitudinal-g-formula-2-steps}). Alternately, we can use a dynamic strategy that uses policies $g_0$ and $g_1$ to assign conditions based on past conditions and observed behavior, using Eqn. \ref{eqn:longitudinal-policy-2-steps} to compute the expected outcome. In the case where the test results $L_0$ and $L_1$ are discrete (\eg pass/fail), the policies $g_0(A_0 \mid L_0)$ and $g_1 (A_1 \mid L_0, A_0, L_1)$ would be conditional probability tables. Different values in these tables would represent different robot behaviors; for example, if the student performed poorly on the first test $(L_0 = 0)$, the robot could be highly likely to offer intervention $(A_0 = 1)$ relative to a robot that is equally likely to offer intervention regardless of the first test score. Overall, the framework described in Section \ref{sec:policy} offers a flexible way to test various hypotheses regarding potential robot tutoring strategies without requiring any experiments to be conducted.


\section{Conducting causal inference studies} \label{sec:pipeline}
\subsection{Illustrating a Causal Study}
In this article, we have provided short case studies that serve as vignettes of how causal methods could be applied to a variety of problems in empirical research. In this section, we further explore the steps that an analyst would take if they wanted to answer a hypothesis using observational data. The analysis is intentionally set up to mimic a randomized experiment, even though no experiment will be conducted, to enhance the interpretability of the results of the hypothetical experiment \cite{hernanUsingBigData2016}. We illustrate this process with an example inspired by the work described in \cite{shiomiRecommendationEffectsSocial2013}, which focuses on social robotics in the context of advertisement. In that work, a robot conversed with people at a shopping mall with the goal of encouraging them to take a coupon, and the primary research question was whether the presence of the robot affected user engagement with the advertising coupon system. Consider a future where such robots are commonly used for advertising and a comprehensive database has been compiled on all aspects of advertising and sales. Suppose that a company has commissioned us to conduct a study on the effectiveness of robot advertising campaigns to justify their larger marketing budgets compared to standard advertising campaigns. In the sections that follow, we will guide the reader through the steps required to apply causal inference in this context, which is summarized in Table \ref{tbl:steps} for reference.

\begin{table}[]
\caption{A guide to conducting causal inference studies}
\begin{tabular}{l|l}
Step & Instruction \\ \hline
1    & Define experimental condition(s) and outcome(s). \\
2    & Define the study population. \\
3 & List all causally relevant variables.\\
4 & Draw a graph.\\
5 & Estimate a causal effect.\\
6 & Consider performing sensitivity analysis.
\end{tabular}
\label{tbl:steps}
\end{table} 




\subsubsection{Define experimental condition(s) and outcome(s)}
To test a hypothesis, we must define the experimental condition(s) and outcome(s) of interest. 
For the robot advertising campaign study, the experimental condition could be a binary variable indicating whether a mall in the database received robot advertising or standard advertising on a particular day. If we have access to more detailed information, such as the type of robot (\eg, anthropomorphic), the experimental condition could be a discrete variable. 
The outcome could be the revenue performance of the mall on that particular day, the results of a customer satisfaction survey, or both. In general, the choice of experimental conditions(s) and outcome(s) depends on the quality of the available data and the research questions that we are attempting to answer.

\subsubsection{Define the study population} Our next step will involve defining the study population whose data we want to analyze. Depending on the question we are asking, we may use only portions of the data that we have access to. In other cases, we may need to refine the question we are asking given the limitations of the data. In our example, we may restrict ourselves to data from shopping malls that the company is interested in (\eg ones they own or have a commercial presence in). We could limit our attention to established malls if we are interested in studying long-term trends. If the company is planning to roll out their advertising campaign in recently opened malls, then we may restrict our attention to newer malls instead.

\subsubsection{List all causally relevant variables}
Since this is an observational study and not an experiment, we must consider all causal variables that could be relevant. These include confounders and mediators (\ie variables that account for the observed relationship between the experimental conditions(s) and the outcome(s)). In the context of the robot advertising campaign, we  consider all confounders that could influence both a mall's sales revenues and whether the mall receives a robot advertising campaign. The socioeconomic context of the mall's location, the number of stores the mall has, its transport accessibility, the types of stores it has, and the amount of foot traffic it gets are all possible confounders. We  also consider mediators, which break down the influence of the advertising campaign on sales revenues into small factors. Examples of potential mediators are the customer's receptivity to robots in public spaces and the behaviors that a robot exhibits to attract customer attention. Finally, there may be variables that come into play after the outcome has already been decided. We can safely exclude these variables because they do not influence the outcome. Regardless of whether a variable is a confounder, mediator, or neither, it can be unobserved. For mediators or variables that take effect after the outcome is decided, this is not a problem, but unobserved variables that are confounders can be a problem depending on their strength. Strong confounders can introduce bias; these should be investigated using sensitivity analysis later (Section \ref{sec:sensitivity}).

\subsubsection{Draw a graph}\label{sec:draw_graph}
Drawing a causal graph helps lay out the causal structure and all of the assumptions involved in our study context. Based on the previous step of listing all causally relevant variables, we include the following variables: let $A$ represent whether robot or standard advertising was used, $C$ represent the socioeconomic context of the mall, $M$ represent the receptivity of the customers to advertising, $Y$ represent the sales revenue, $G$ represent the public perception of the quality of the mall, and $Z$ be general economic conditions. Given these variables, we can use domain knowledge to draw a graph. The use of robot advertising (versus standard advertising) is reasonably conjectured to affect the sales revenue outcome, which leads us to draw an edge $A \to Y$. The socioeconomic context of the mall is likely to  influence both whether companies deploy robots there as well as the overall revenue levels, resulting in the edges $C \to A$ and $C \to Y$. We might consider the receptiveness to advertising generally as $M$; it is influenced by the type of advertising used and the socioeconomic context of the mall and in turn influences both sales revenue and public perception, resulting in the edges $A \to M, C \to M, M \to G,$ and $M \to Y$. The public perception of the quality of the mall is influenced by all of the other variables, resulting in edges $C \to G, A \to G, M \to G,$ and $Y\to G$. Finally, we may posit that general economic conditions influence both $C$ and $Y$, so we insert $ C\leftarrow Z \rightarrow Y$.

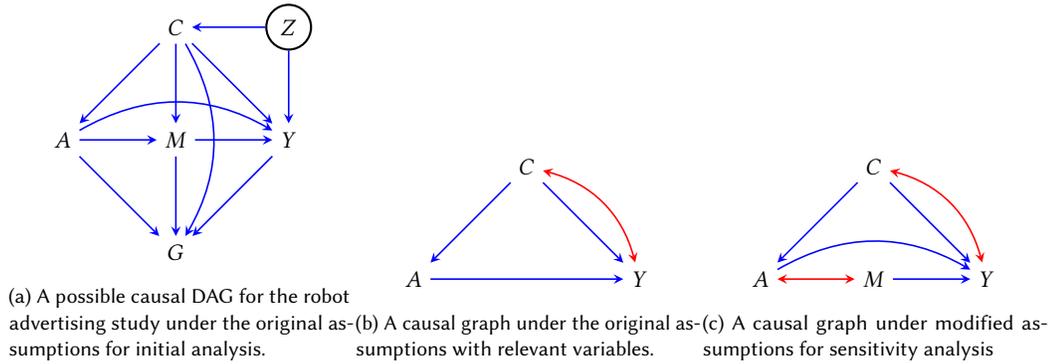
\begin{figure}[t]
    \begin{subfigure}{0.3\textwidth}
    \centering
    \begin{tikzpicture}[
        > = stealth, 
        auto,
        node distance = 1.5cm, 
        semithick 
        ]

        \tikzstyle{state}=[
        draw = none,
        fill = white,
        minimum size = 2mm
        ]
        \tikzstyle{fixed}=[
        draw=black,
        rectangle, 
        thick,
        fill = white,
        minimum size = 2mm
        ]
        \tikzstyle{hidden}=[
        draw=black,
        circle, 
        thick,
        fill = white,
        minimum size = 2mm
        ]

        \node[state] (A) {$A$};
        \node[state] (M) [right of=A]{$M$};
        \node[state] (C) [above of=M]{$C$};
        \node[state] (G) [below of=M]{$G$};
        \node[state] (Y) [right of=M]{$Y$};
        \node[hidden] (Z) [above of=Y]{$Z$};

        \path[->, blue] (A) edge node {} (M);
        \path[->, blue] (M) edge node {} (Y);
        \path[->, blue] (A) edge node {} (G);
        \path[->, blue] (Y) edge node {} (G);
        \path[->, blue] (C) edge node {} (A);
        \path[->, blue] (C) edge node {} (Y);
        \path[->, blue] (C) edge node {} (M);
        \path[->, blue] (M) edge node {} (G);
        \path[->, blue, bend left=30] (A) edge node {} (Y);
        \path[->, blue, bend left=30] (C) edge node {} (G);
        \path[->, blue] (Z) edge node {} (Y);
        \path[->, blue] (Z) edge node {} (C);
        %
    \end{tikzpicture}
    \subcaption{A possible causal DAG for the robot advertising study under the original assumptions for initial analysis.}
    \label{fig:example_graph}
    \end{subfigure}
    \begin{subfigure}{0.3\textwidth}
    \centering
    \begin{tikzpicture}[
        > = stealth, 
        auto,
        node distance = 1.5cm, 
        semithick 
        ]

        \tikzstyle{state}=[
        draw = none,
        fill = white,
        minimum size = 2mm
        ]
        \tikzstyle{fixed}=[
        draw=black,
        rectangle, 
        thick,
        fill = white,
        minimum size = 2mm
        ]
        \tikzstyle{hidden}=[
        draw=black,
        circle, 
        thick,
        fill = white,
        minimum size = 2mm
        ]

        \node[state] (A) {$A$};
        \node[state] (C) [above of=M]{$C$};
        \node[state] (Y) [right of=M]{$Y$};

        \path[->, blue] (A) edge node {} (Y);
        \path[->, blue] (C) edge node {} (Y);
        \path[->, blue] (C) edge node {} (A);
        \path[<->, red, bend left=30] (C) edge node {} (Y);
        %
    \end{tikzpicture}
    \subcaption{A causal graph under the original assumptions with relevant variables.}
    \label{fig:example_graph_relevant}
    \end{subfigure}
    \begin{subfigure}{0.3\textwidth}
    \centering
    \begin{tikzpicture}[
        > = stealth, 
        auto,
        node distance = 1.5cm, 
        semithick 
        ]

        \tikzstyle{state}=[
        draw = none,
        fill = white,
        minimum size = 2mm
        ]
        \tikzstyle{fixed}=[
        draw=black,
        rectangle, 
        thick,
        fill = white,
        minimum size = 2mm
        ]
        \tikzstyle{hidden}=[
        draw=black,
        circle, 
        thick,
        fill = white,
        minimum size = 2mm
        ]

        \node[state] (A) {$A$};
        \node[state] (M) [right of=A] {$M$};
        \node[state] (C) [above of=M]{$C$};
        \node[state] (Y) [right of=M]{$Y$};

        \path[<->, red] (A) edge node {} (M);
        \path[->, blue] (C) edge node {} (Y);
        \path[->, blue] (C) edge node {} (A);
        \path[->, blue] (M) edge node {} (Y);
        
        \path[<->, red, bend left=30] (C) edge node {} (Y);
        \path[->, blue, bend left=30] (A) edge node {} (Y);
        %
    \end{tikzpicture}
    \subcaption{A causal graph under modified assumptions for sensitivity analysis}
    \label{fig:example_graph_modified}
    \end{subfigure}
    \caption{Possible causal graphs for the causal study. Let $A$ denote whether robot or standard advertising was used, $C$ denote the socioeconomic status of the mall, $M$ denote the receptivity of the customers to advertising, $Y$ denote sales revenue, and $G$ denote the public perception of the quality of the mall.}
    \label{fig:example_graph_sets}
\end{figure}

The resulting graph is shown in Figure \ref{fig:example_graph}. Note that the edges we drew may be subject to uncertainty, which we discuss in subsequent sensitivity analysis (Section \ref{sec:sensitivity}). 

\subsubsection{Estimate a causal effect.}
Based on our hypothesis and the graph we drew, we can decide on the causal effect we want to measure and estimate it. For our study, we may choose to estimate the average causal effect $\E[Y(a)]$. We can estimate $\E[Y(a)]$ using the estimators detailed in Equations \ref{eqn:g-formula-est} or \ref{eqn:ipw-est} from Section \ref{sec:observed-common-causes}, ignoring variables $M$ and $G$ and considering $C$ a confounding variable. The relevant variables are represented in \cref{fig:example_graph_relevant} and the identifying functional is given as 
\[\E[Y(a)] = \sum_C \E[Y \mid A=a, C]p(C)\]
with backdoor adjustment set $C$. Estimators for this quantity include the g-formula and inverse probability weights.

\subsubsection{Consider performing sensitivity analysis}
\label{sec:sensitivity}
Sensitivity analysis allows us to check how sensitive our inferences are to our initial assumptions. Because causal inference cannot be validated without conducting an experiment, these analyses are important to establish trust in the results. 
As an example, consider the following changes to the assumptions made in \cref{sec:draw_graph}.
First, perhaps we believe that the relationship between the advertising type $A$  and receptivity $M$ is not a directed causal relationship, but simply a correlation represented by a bidirected edge.
Second, we know that the variables $M$ and $G$ (corresponding to receptivity and perception) are most difficult to measure, since these would be collected via a survey or questionnaire conducted by an interviewer. If we believe that these variables are collected with error due to subjective bias, then considering measurement error methods may be of interest.

Under these new assumptions, the graph can be represented as \cref{fig:example_graph_modified}. Applying the methods discussed from \cref{sec:observed-common-causes}, the backdoor adjustment set is $\{C, M\}$ and the identifying functional is given as 

\[\E[Y(a)] = \sum_{C, M} \E[Y \mid A=a, C, M] p(C, M).\]

 
%

\subsection{Useful Software}

We direct the reader to the code that accompanies this article at \url{https://gitlab.com/causal/causal_hri}. The repository consists of a Python package that implements many of the methods discussed in the article from scratch and also includes several Jupyter notebooks that demonstrate how these methods are used on simulated data.

The causal inference software ecosystem is currently not as centralized as the machine learning software ecosystem where users have, for example, coalesced around a few standard packages such as \texttt{scipy}, \texttt{sklearn}, \texttt{statsmodels}, or deep learning frameworks. Instead, the software ecosystem is highly fragmented, with many packages existing for a small subset of functions. These packages are developed mostly for the \texttt{R} programming language and not the \texttt{python} language that has become the lingua franca of machine learning. It is not uncommon for causal inference practitioners to re-implement existing methods when conducting their own analysis. We list some of the software resources that are currently available for causal inference below:

\begin{itemize}
    \item
For a general reference, \cite{hernanCausalInferenceWhat2020} provides causal inference exercises, a large portion of which include code examples in \texttt{python} and \texttt{R}. This includes observed confounding in both static and longitudinal settings. 

\item For scenarios with observed confounding only (see Section \ref{sec:observed-common-causes}), \cite{linGfoRmulaPackageEstimating2019} provide a g-formula estimator (corresponding to Equation \ref{eqn:g-formula}) implemented in \texttt{R}.

\item For causal discovery (Section \ref{sec:causal-discovery}), a few well-established packages exist. \texttt{pcalg} \cite{kalischCausalInferenceUsing2012,hauserCharacterizationGreedyLearning2012}  is an \texttt{R} package, while \texttt{tetrad} is a \texttt{java} package. Both implement popular causal discovery algorithms such as PC, GES, and FCI, all of which learn causal graphs.

\item For general graph drawing and identification of causal effects, we recommend \texttt{ananke} \cite{leeAnanke2020}.
\end{itemize}
\section{Discussion} \label{sec:conclusion}


\subsection{Key Takeaways}
The primary goal of this article is to convey to readers the possibilities that exist for doing hypothesis-based science in the absence of fully randomized experiments. While we do not cover the complete array of tools that a practitioner hoping to apply these methods will require, we hope this article can serve as a starting point for understanding HRI problems through a causal inference framework. We summarize some of the key takeaways from this article:

\subsubsection{Designing experiments that enable valid causal inference}
The first takeaway is intended for those researchers who conduct experiments. These readers should pay attention to their experiments to ensure that the conditions required for associational methods to enable causal inference in fact hold (\eg, consistency and conditional ignorability) (see Sections \ref{sec:randomized-experiments} and \ref{sec:observed-common-causes}). Recall the example from Section \ref{sec:existing-methods-fail} where a sequentially randomized experiment could lead to erroneous inference when associational methods such as linear regressions are used to analyze study data. Additionally, recall our discussion of transportability in Section \ref{sec:transportability}, which involved the experiment being conducted and the question being asked existing in different domains. In both of these examples, performing causal inference using experiments would be invalid.

\subsubsection{Taking a causal inference perspective}
The second takeaway is to think about problems with causality in mind. One easy way to start is by using causal graphical models, such as DAGs. These models are intuitive and can help with quickly summarizing the assumptions that have been made about a problem's dynamics. Having a clear picture of the problem at hand can be helpful, even if the analyst does not proceed to use this graphical model in more advanced ways. 

\subsubsection{Collecting data that enables valid causal inference}
The third takeaway is that even when observational studies are being conducted without a causal question in mind, attention should be paid to aspects that would enable data to be analyzed in a causal manner. In particular, researchers should think about potential confounders in the study and include them in data collection if possible. This can increase the usefulness of the dataset for future analysis. In general, this means researchers may need to collect more information about the study context and participant demographics.

\subsection{Limitations} 
Causal inference methods can be useful tools for studying HRI when running randomized experiments is not feasible, but they are not without limitations. We emphasize that any causal result is valid only if certain assumptions hold. These assumptions, such as consistency or conditional ignorability, are not testable unless even stronger assumptions are made. Thus, the validity of a result is only as strong as the researcher's belief in the assumptions. Careful examination and scrutiny of assumptions, aided by sensitivity analysis to check how much influence those assumptions wield, is important for valid causal inference. Furthermore, not all questions seek a quantitative answer, and causal inference does not seek to displace qualitative lines of inquiry. Many scenarios exist where the relevant variables are not suited to quantitative measurement or quantitative research is impractical due to small sample sizes, and there are valid critiques of quantitative methods when they are deployed at the exclusion of cultural, social, and societal aspects of HRI \cite{fischerEffectConfirmedPatient2021,seibtComplexityHumanSocial2021}. We also note that causal methods are but one among several methods for increasing the ecological validity of HRI research; behavioral scientists have proposed various methods, such as peer-reviewed data analysis sessions and stricter specification of contexts of interest in publications, that can improve the applicability of human interaction research to real-world settings \cite{albert2018improving, holleman2020real}. We hope that the methods presented in this paper complement other lines of inquiry and help researchers who make causal claims from quantitative work do so in a rigorous framework. 

\subsection{Comparing Causal and Statistical Methods}
Some readers may wonder how causal inference and statistical (non-causal) inference are related. In causal inference, analysts ask questions about data that they don't technically have. For example, we might collect some observational data about robot behavior policies in robots, but then wonder which policy is best had we deployed and collected data from a randomized trial. Importantly, we did not actually perform the randomized trial and so have no data from it. Consequently, in order for the causal conclusions to have any basis in reality, the analyst must consider explicitly how the desired parameter is a function of the observed data distribution. This question is often left implicit in most statistical inference problems as the questions tend to focus on the data at hand. This fundamental difference in approaches leads to a divergence in the two fields, and to explain the nuances of this relationship we will turn to a concrete example.

Consider a simple example (\eg \cref{sec:observed-common-causes}) where we have some binary treatment/condition A, baseline covariates, and study outcome Y. We consider potential outcomes $Y(0)$ and $Y(1)$, which represent random variables if $A$ were set to 0 and 1, respectively. Abstractly, we might consider what is called the ``full data distribution'' $p_\textrm{full} (Y(1), Y(0), A,C)$, which is a distribution over both potential outcomes, the treatment, and the covariates. Having this distribution represents the scenario in which we get to observe each participant under both of the treatments, which is in practice impossible. What we observe in practice is called the observed data distribution $p_\textrm{obs}(Y, A, C)$ where $Y = Y(1) * A + Y(0) * (1 - A)$. Here, Y is the potential outcome corresponding to the observed treatment level $A$, and so $p_\textrm{obs}$ is a coarsened version of $p_\textrm{full}$.

Causal inference involves some target parameter of interest (\eg the average causal effect $\E[Y(1) - Y(0)]$ that represents the result of a randomized experiment) that is in $p_\textrm{full}$, which is to be expressed using data from $p_\textrm{obs}$. To do this, we need some assumptions that allow us to ``invert'' this coarsening process. These assumptions are a causal model, and are often represented using causal graphs as depicted in the article. Using these assumptions, we can attempt to express the target parameter as a functional of the observed data $p_\textrm{obs}$, in a process called ``identification''. It is possible that the process of identification fails, because the target parameter cannot be identified from the given data and assumptions. At this point the analyst can either collect more variables, or introduce further assumptions to continue.

Note that all causal interpretation rests inside the functional. Once obtained by the analyst, strictly speaking we may consider the functional to be a function of data devoid of causal interpretation, and use any tools in the statistical inference literature to estimate this function, in a step called ``estimation''. In this article we provided frequentist parametric approaches through the inverse probability weighting and g-formula methods in \cref{sec:observed-common-causes}, but other approaches such as frequentist semi-parametric methods \cite{tsiatisSemiparametricTheoryMissing2006} or Bayesian methods like Bayesian additive regression trees \cite{chipmanBARTBayesianAdditive2010} also exist. In choosing a method there are various trade-offs one makes, such as the implementation complexity of the estimator, its statistical efficiency properties (if any), and the assumptions that must hold in order for the estimator to be valid. Another approach which we do not explore in this article is to be fully Bayesian about the full data parameter and leave issues regarding identifiability to the posterior distribution, in effect combining the identification and estimation steps \cite{baldiBayesianCausality2020,rubinCausalInferenceUsing2005}.

\subsection{Further Readings}
To the authors' knowledge, this is the first work on causal inference specifically written for observational studies in an HRI context. However, there have been a variety of causal inference works written for a beginner audience, assuming only a basic familiarity with probability and statistics. We list some examples of such works that can serve as additional references for HRI researchers:

\begin{itemize}
\item \cite{hernanCausalInferenceWhat2020} is a recent causal inference textbook freely available online. The book covers many key topics of causal inference such as defining the causal question (using nonparametric identification), basic to advanced estimation strategies, and causal inference over longitudinal settings. This material is covered assuming only an undergraduate-level of probability and statistics. In particular, Part I of this textbook contains some useful exposition on the core concepts of causal inference. Part II covers strategies for estimation once a causal effect has been identified. This becomes particularly important when real data is involved. 

\item \cite{pearlCausality2009} is another standard causal inference textbook in the field. This book builds explicitly on the causal graphical model, although there are some minor notational differences (\eg notations for causal effects are phrased in terms of the $\doo$-operator instead of the equivalent potential outcome). However, the graph-centric view may be useful for readers interested in formalizing their problems using graphical models.
\end{itemize}

\section{Conclusion}
In this article, we introduced causal inference methods for enabling quantitative HRI research using observational data. We highlighted the need for such methods by showing how recent work can be extended through field studies and demonstrated how causal inference enables hypotheses to be answered using field or observational studies through simple HRI examples and case studies. We believe that these methods can be useful to HRI researchers who seek additional rigor and mathematical precision in defining scientific hypotheses, as well as researchers who wish to keep HRI in the field to maximize external validity. We hope that these methods and their possibilities will excite researchers and expand the field studies that are being conducted in HRI.

\begin{acks}
G.A. acknowledges support in part by the National Science Foundation Graduate Research Fellowship Program under Grant No. DGE-1746891 and the Malone Center for Engineering in Healthcare at the Johns Hopkins University.

J.J.R.L. would like to thank Amanda Hilliard at the JHU Center for Leadership Education for editorial and writing assistance.
\end{acks}



\bibliographystyle{ACM-Reference-Format}
\bibliography{references, library}

\end{document}